\documentclass[letterpaper]{article}
\usepackage{aaai24}

\usepackage{times}  
\usepackage{helvet}  
\usepackage{courier}  
\usepackage[hyphens]{url}  
\usepackage{graphicx} 
\usepackage{natbib}  
\usepackage{caption} 
\usepackage{times}
\usepackage{amsmath}
\usepackage{cases}
\usepackage{enumitem}
\usepackage{float}

\usepackage{tabularx}
\usepackage{booktabs}
\usepackage{amssymb}
\usepackage{color}  
\usepackage{subfig}
\usepackage{marvosym}

\begin{document}
%
\title{A Fully Data-Driven Approach for Realistic Traffic Signal Control Using\\ Offline Reinforcement Learning}
\author {
    Jianxiong Li\textsuperscript{\rm 1,2}*,
    Shichao Lin\textsuperscript{\rm 3}*,
    Tianyu Shi\textsuperscript{\rm 4},
    Chujie Tian\textsuperscript{\rm 5},
    Yu Mei\textsuperscript{\rm 5},
    Jian Song\textsuperscript{\rm 5},\\
    Xianyuan Zhan\textsuperscript{{\rm 1,6}\Letter},
    Ruimin Li\textsuperscript{{\rm 3}\Letter},
}
\affiliations {
    \textsuperscript{\rm 1} Institute for AI Industry Research (AIR), Tsinghua University, Beijing, China\\
    \textsuperscript{\rm 2} School of Vehicle and Mobility, Tsinghua University, Beijing, China\\
    \textsuperscript{\rm 3} Department of Civil Engineering, Tsinghua University, Beijing, China\\
    \textsuperscript{\rm 4} Department of Civil and Mineral Engineering, University of Toronto, Toronto, Canada\\
    \textsuperscript{\rm 5} Baidu Inc., Beijing China\\
    \textsuperscript{\rm 6} Shanghai Artificial Intelligence Laboratory, Shanghai, China\\
    \{li-jx21, linsc19\}@mails.tsinghua.edu.cn, zhanxianyuan@air.tsinghua.edu.cn, lrmin@tsinghua.edu.cn
}
\maketitle
\begin{abstract}
The optimization of traffic signal control (TSC) is critical for an efficient transportation system. In recent years, reinforcement learning (RL) techniques have emerged as a popular approach for TSC and show promising results for highly adaptive control. However, existing RL-based methods suffer from notably poor real-world applicability and hardly have any successful deployments. The reasons for such failures are mostly due to the reliance on over-idealized traffic simulators for policy optimization, as well as using unrealistic fine-grained state observations and reward signals that are not directly obtainable from real-world sensors. In this paper, we propose a fully \underline{D}ata-\underline{D}riven and simulator-free framework for realistic \underline{T}raffic \underline{S}ignal \underline{C}ontrol (D2TSC). Specifically, we combine well-established traffic flow theory with machine learning to construct a reward inference model to infer the reward signals from coarse-grained traffic data. With the inferred rewards, we further propose a sample-efficient offline RL method to enable direct signal control policy learning from historical offline datasets of real-world intersections. To evaluate our approach, we collect historical traffic data from a real-world intersection, and develop a highly customized simulation environment that strictly follows real data characteristics. We demonstrate through extensive experiments that our approach achieves superior performance over conventional and offline RL baselines, and also enjoys much better real-world applicability.

\end{abstract}

\section{Introduction}
Traffic signal control (TSC) is an important and challenging real-world problem, which is central to urban congestion alleviation and improving traffic system efficiency. Over the years, the traffic signal control problem has attracted considerable attention from research communities. Although widely used in practice, conventional transportation engineering methods for TSC~\citep{webster1958traffic,hunt1982scoot,lowrie1990scats,gartner1991multi,cools2013self,yan2019network,lu2023optimization} heavily rely on domain knowledge (such as pre-defined rules) and field calibration, which are not flexible enough to handle highly dynamic traffic conditions and can be costly for large-scale implementation. On the other hand, the recently emerged reinforcement learning (RL) based TSC methods enable direct policy learning from environment interactions without making strong assumptions about the traffic, thus holding great promise for achieving general-purpose and fully adaptive traffic signal control~\citep{wei2018intellilight,zheng2019diagnosing,zang2020metalight,wei2021recent}. However, while RL-based methods have achieved impressive performance in traffic simulation environments, there have been few successful attempts to deploy them in the real world.

The poor real-world applicability of existing RL-based TSC methods primarily stems from two notable issues.
First, real-world intersections are complicated open systems with many influencing factors, such as heterogeneous driving and pedestrian behaviors, making realistic simulation difficult. Most RL-based TSC methods are built upon conventional online RL framework, which heavily relies on extensive interactions in idealized traffic simulations, and suffers from severe sim-to-real transfer issues~\citep{zhang2023data}. Second, the data collected from real-world intersections are typically coarse-grained, providing much less information as compared to the full-state observation obtainable from traffic simulators. For example, many existing RL-based models leverage highly informative state features from simulators for signal control, such as the position of vehicles~\citep{van2016coordinated,wei2018intellilight,liang2019deep, mo2022cvlight} and queue lengths~\citep{li2016traffic, zheng2019diagnosing,zhang2019cityflow, yoon2021transferable}, however, such information is typically not directly obtainable from the widely used detectors for real-world traffic control systems~\citep{hunt1982scoot,sims1980sydney}.

The above issues necessitate the need for developing a completely data-driven TSC framework to bypass the limitations of traffic simulators and improve deployment feasibility. The recently emerged offline RL~\citep{fujimoto2019off,levine2020offline} provides an attractive paradigm for RL-based TSC policy learning using historical data from real-world intersections. However, adopting an offline RL solution for TSC also faces two major challenges. First, the fine-grained reward signal is hard to obtain from real signalized intersections. Important evaluation metrics such as queue lengths and delays, although easily obtainable in simulation, are not monitored nor collected in real-world signal control data, providing no reward information for data-driven optimization. Second, the amount of data we can obtain from real signalized intersections may be very limited. For instance, if we sample in 5-minute intervals, a whole month's historical traffic data of an intersection will only correspond to about 8,600 state-action samples, much smaller in dataset size as compared to typical offline RL benchmark tasks~\citep{fu2020d4rl}.

To tackle the above challenges, we develop a fully \underline{D}ata-\underline{D}riven framework for real-world \underline{T}raffic \underline{S}ignal \underline{C}ontrol (D2TSC). Our framework combines the merits of both well-established traffic flow theories and a state-of-the-art offline RL algorithm to achieve sample-efficient and deployment-friendly TSC policy learning. Specifically, we first develop a data-driven reward inference model based on shockwave theory~\citep{lighthill1955kinematic,richards1956shock,daganzo1997fundamentals,jin2015point} and Gaussian process interpolation process using realistic coarse-grained intersection data. With the learned reward, we develop an in-sample learning offline RL method with customized state and action encoding design according to real-world TSC data characteristics, as well as data augmentation for sample-efficient policy learning.

To evaluate our framework, we collect real-world data from a signalized intersection in China and build a highly customized simulation environment with state observations strictly following the real-world detection (e.g., 5-minute traffic flow and spatial vehicle count of each lane in the 150m range every 5 seconds monitored by license-plate recognition cameras). We generate 3 months of data based on the actual traffic flows and timing plans and train all models based on these limited data. Numerical experiments show that our framework outperforms all other baselines and exhibits very good deployability, given its sample efficiency, compatibility with coarse-grained TSC data, and capability to infer reward signals directly from the data.


\section{Related Work}
\subsection{Traffic Signal Control Using RL}
RL methods allow direct learning of an optimized policy through interactions with the unknown environment. This nice property has long attracted researchers to apply RL to solve traffic signal control problems. Early works~\citep{wiering2004intelligent,cai2009adaptive,abdulhai2003reinforcement} use the tabular Q-learning method to solve highly simplified problems, with states required to be discrete and low-dimensional. With the development of deep RL, many recent works leverage highly expressive deep neural networks to model more complex and information-rich state inputs to improve TSC performance~\citep{wei2018intellilight,zheng2019learning,zheng2019diagnosing,rodrigues2019towards,oroojlooy2020attendlight,wei2021recent,shabestary2022adaptive}. Other advanced modeling tools have also been adopted to improve the signal control scalability and robustness. For example, multi-agent RL and graph neural networks have been adopted to scale the TSC control from a single intersection to multiple intersections in a road network ~\citep{wei2019colight,iqbal2019actor,chu2019multi,chen2020toward,devailly2021ig}; meta-learning has also been explored for RL-based TSC~\citep{zang2020metalight}, to help the learned model adapt quickly and stably in new traffic scenarios. 

Although the aforementioned methods achieve impressive performance in simulation environments, they are still far from being able to deploy in real-world scenarios. Their poor practical feasibility mainly stemmed from the over-reliance on the traffic simulation environments, as analyzed in our Introduction section. Recently, there have been some attempts~\citep{kunjir2022offline, zhang2023data} to adopt offline RL methods to solve the TSC problem, which allows policy learning from pre-collected offline datasets. However, these studies still use unrealistic fine-grained state inputs and ground-truth reward information that are only obtainable from simulation environments for modeling, completely deviating from their original intention of offline learning. In this study, we address the drawbacks of the previous works by proposing a novel offline RL framework for traffic signal control, which enables completely simulator-free and deployment-friendly policy learning using realistic TSC data.

\subsection{Offline Reinforcement Learning}
Reinforcement learning provides a promising way to solve TSC problems. However, existing studies primarily adopt \textit{online RL} methods, in which the control policy is learned via interacting with the environment following a trial-and-error paradigm. However, the requirement of online interactions in online RL inhibits the successful applications of RL methods in many complex real-world tasks, as interaction with the real system during policy learning can be costly, unsafe, or ethically problematic, and high-fidelity simulators are also hard to build~\citep{levine2020offline,kiran2021deep,zhan2021deepthermal}.
On the contrary, the recently emerged \textit{offline RL} methods smartly tackle the challenges of online RL via optimizing policies using only pre-collected offline datasets without further online interactions~\citep{fujimoto2019off, wu2019behavior, kumar2020conservative}. 
Therefore, offline RL holds great promise to utilize existing historical operational data recorded in existing TSC control systems to facilitate signal control optimization.

However, the absence of online interactions also poses new technical difficulties for offline policy learning. Concretely, directly applying online RL methods in the offline setting faces significant training instability when performing model evaluation on samples outside the dataset distribution (also referred to as out-of-distribution (OOD)), where estimation errors can quickly build up and cause the issue of distributional shift and severe value overestimation~\citep{fujimoto2019off, kumar2020conservative}. 
One straightforward approach to address these problems is through \textit{policy constraints}, which incorporates a constraint into the policy learning to prevent excessive deviation of the optimized policy from the behavior policy (the policy present in the offline dataset)~\citep{fujimoto2019off, wu2019behavior, li2023when}. There are also some \textit{value regularization} methods that penalize the value function to assign low values at OOD regions to mitigate overestimation errors~\citep{kumar2020conservative, bai2021pessimistic, an2021uncertainty, niu2022trust}.
Although effective, these methods still need to evaluate the value function on policy-induced, possibly OOD actions, which can induce potential instability. Recently, \textit{in-sample learning} has emerged as a promising alternative for training offline RL without the need to query values on OOD samples~\citep{sql, iql, xu2022policy, garg2023extreme}, thereby effectively addressing the issues of overestimation errors in OOD regions. In our study, we also adopt the in-sample learning offline RL framework, with customized state-action encoding designs and a sample-efficient data augmentation scheme tailored to realistic TSC optimization problems. 

\section{Preliminary} \label{sec:2}
\subsection{Markov Decision Process}
The RL problem is typically formulated as a Markov Decision Process (MDP), modeled by a tuple $\langle\mathcal{S},\mathcal{A},r,\gamma,\mathcal{P},\rho\rangle$. $\mathcal{S}$ and $\mathcal{A}$ denote the state and action space. $r: \mathcal{S}\times\mathcal{A}\rightarrow \mathbb{R}$ is the reward function. $\gamma\in(0, 1)$ denotes the discount factor.  $\mathcal{P}: \mathcal{S}\times\mathcal{A}\rightarrow\mathcal{S}$ represents the transition dynamic and $\rho$ is the initial state distribution. The goal of RL is to find an optimal policy $\pi^*: \mathcal{S}\rightarrow\mathcal{A}$ that can maximize the expected discounted cumulative rewards:
\begin{equation}
    \max_{\pi} \mathbb{E}\left[\sum_{t=0}^\infty \gamma^t r(s_t,a_t)|s_0\sim\rho, a_t\sim\pi, s_{t+1}\sim \mathcal{P}\right].
    \label{equ:rl_objective}
\end{equation}
The expected discounted cumulative reward presented in Eq.~(\ref{equ:rl_objective}) is typically expressed as a state-value function $V^\pi(s):=\mathbb{E}\left[\sum_{t=0}^\infty\gamma^tr(s_t,a_t)|s_0=s,a_t\sim\pi,s_{t+1}\sim\mathcal{P}\right]$ or an action-value function $Q^\pi(s,a):=\mathbb{E}[\sum_{t=0}^\infty\gamma^tr(s_t,a_t)|$ $s_0=s,a_0=0,a_{t+1}\sim\pi,s_{t+1}\sim\mathcal{P}]$. 
In practice, modern deep RL methods typically approximate the action-value function $Q^\pi(s,a)$ using deep neural networks by
minimizing the squared Bellman error~\citep{lillicrap2015continuous, mnih2016asynchronous, fujimoto2018addressing, haarnoja2018soft}:
\begin{equation}
    \begin{aligned}
    Q^\pi=\arg\min_Q \mathbb{E}_{(s,a,r,s')\sim\mathcal{D}}[(r(s,a)\\
    +\gamma \mathbb{E}_{a'\sim\pi(\cdot|s')}Q(s',a')-Q(s,a))^2] ,\label{equ:min_Q}
    \end{aligned}
\end{equation}

where, $\mathcal{D}$ is a data buffer (also known as replay buffer) that contains historical transitions $\mathcal{D}:=\{(s,a,s',r)_i\}$ gradually filled during online interactions, or a fixed offline dataset in the offline RL setting.
With the estimated value function,
most RL methods learn an optimized policy $\pi(a|s)$ by maximizing the value function as:
\begin{align}
    \max_\pi \mathbb{E}_{s\sim\mathcal{D},a\sim\pi(\cdot|s)}\left[Q^\pi(s,a)\right].\label{equ:min_pi}
\end{align}
By repeatedly alternating between Eq.~(\ref{equ:min_Q}-\ref{equ:min_pi}), the reward maximization objective stated in Eq.~(\ref{equ:rl_objective}) can be approximately solved~\citep{lillicrap2015continuous, fujimoto2018addressing, haarnoja2018soft,mnih2016asynchronous}.





\subsection{Offline Reinforcement Learning}

Under the offline RL setting, the dataset $\mathcal{D}$ introduced above is fixed and is pre-collected by some unknown behavior policies $\mu(a|s)$ without the possibility of further interacting with the environment. In this case, directly adopting online RL methods in Eq.~(\ref{equ:min_Q}-\ref{equ:min_pi}) will lead to severe instability caused by distributional shift~\citep{fujimoto2019off} and approximation error accumulations~\citep{kumar2020conservative}. Specifically, maximizing the action value in Eq.~(\ref{equ:min_pi}) may cause the learned policy to deviate from the offline data distribution to some OOD regions. This will introduce approximation errors when learning the action-value function in Eq.~(\ref{equ:min_Q}). Such errors will quickly build up and cause severe value overestimation if no regularization is used to stabilize the training, leading to training instability and policy learning failures~\citep{fujimoto2019off, kumar2020conservative, wu2019behavior, xu2021offline}.


To combat these challenges, the most straightforward way is to regularize the optimized policy to stay close to the data distribution~\citep{fujimoto2019off, kumar2020conservative, sql, wu2019behavior}. By doing so, the maximization over the value function in Eq.~(\ref{equ:min_pi}) is performed within the offline data distribution, thus avoiding value overestimation at OOD regions. A neat choice to achieve this is to augment the standard RL objective in Eq.~(\ref{equ:rl_objective}) with a behavior regularization term and solves a \textit{behavior regularized} MDP~\citep{sql}:
\begin{equation}
    \max_{\pi} \mathbb{E}\left[\sum_{t=0}^\infty \gamma^t \left(r(s_t,a_t)-\alpha f\left(\frac{\pi(a_t|s_t)}{\mu(a_t|s_t)}\right)\right)\right],
    \label{equ:brMDP}
\end{equation}
where $f(\cdot)$ can be any $f$-function from $f$-divergences. 
This objective aims to maximize the cumulative return while minimizing the deviation of the optimized policy $\pi$ to the behavior policy $\mu$ by minimizing the $f$-divergence regularization term $f(\pi/\mu)$, thereby ensuring the learned policy stays close to the data distribution and avoids the distributional shift issue.
Akin to the treatment in max-entropy RL~\citep{haarnoja2017reinforcement, haarnoja2018soft}, the most direct approach to solving the behavior regularized MDP is to augment a behavior regularization term $f(\pi/\mu)$ upon the original objectives in Eq.~(\ref{equ:min_Q}) and Eq.~(\ref{equ:min_pi}):
\begin{equation}
    \begin{aligned}
        \min_Q\mathbb{E}_{(s,a,s')\sim\mathcal{D}}\Bigg[\Bigg(&r(s,a)+\gamma \mathbb{E}_{a'\sim\pi(\cdot|s')}\Bigg[Q(s', a')\\
        &-\alpha f\Bigg(\frac{\pi(a'|s')}{\mu(a'|s')}\Bigg)\Bigg]-Q(s,a)\Bigg)^2\Bigg],\label{equ:ac_q}\\
    \end{aligned}
\end{equation}

\begin{align}
\max_\pi\mathbb{E}_{s\sim\mathcal{D},a\sim\pi}\left[Q(s,a)-\alpha f\left(\frac{\pi(a|s)}{\mu(a|s)}\right)\right].\label{equ:ac_pi}
\end{align}

Intuitively, the augmented action-value learning step in Eq.~(\ref{equ:ac_q}) penalizes the action-value $Q(s,a)$ at the regions that undergo large distributional shift measured by the $f$-divergence. Meanwhile, the augmented policy learning step in Eq.~(\ref{equ:ac_pi}) seeks to maximize the value function and force the policy to stay close to the behavior policy using the behavior regularizer. More importantly, \cite{sql} recently demonstrates that the objectives in Eq.~(\ref{equ:ac_q}-\ref{equ:ac_pi}) can equivalently transfer to a broad class of SOTA in-sample learning offline RL methods~\citep{sql, iql, garg2023extreme}, which enjoy great training stability and SOTA performances. In this paper, we also resort to the in-sample version of Eq.~(\ref{equ:ac_q}-\ref{equ:ac_pi}) for its superior performances (Section~\ref{sec:in_sample}).

\subsection{Traffic Signal Control Using Real-World Data}
Before introducing our framework, we first provide the terminologies and notations in TSC optimization:
\begin{itemize}[leftmargin=*,topsep=0pt,noitemsep]
    \item \textbf{Traffic movement}: A traffic movement is defined as traffic moving across the intersection towards a certain direction, such as left-turn, straight, and right-turn. 
    \item \textbf{Signal phase} and \textbf{phase order}: a signal phase $p$ is a continuous period during which the traffic signal for a specific traffic movement displays the same condition (red or green). For example, the activation of the "North-South Straight" phase illuminates green signals for the southbound and northbound straight lanes, while other entrance lanes show red signals. We denote $\mathcal{P}$ as the set of all phases at an intersection.
    This paper focuses on the main signal phases, ignoring the transition phases like yellow and all-red. A phase order (a.k.a. phase structure) refers to a sequence consisting of specific signal phases that are activated sequentially and cyclically. 
    \item \textbf{Signal timing plan}: signal timing plan is a collection of parameters and logic to allocate the right-of-way at a signalized intersection. It specifies the signal cycle length $T_c$ and the green $T_g^p$ and red times $T_r^p$ of each phase $p\in\mathcal{P}$.
\end{itemize}

\begin{figure*}[t]
    \centering
    \includegraphics[width=0.95\textwidth]{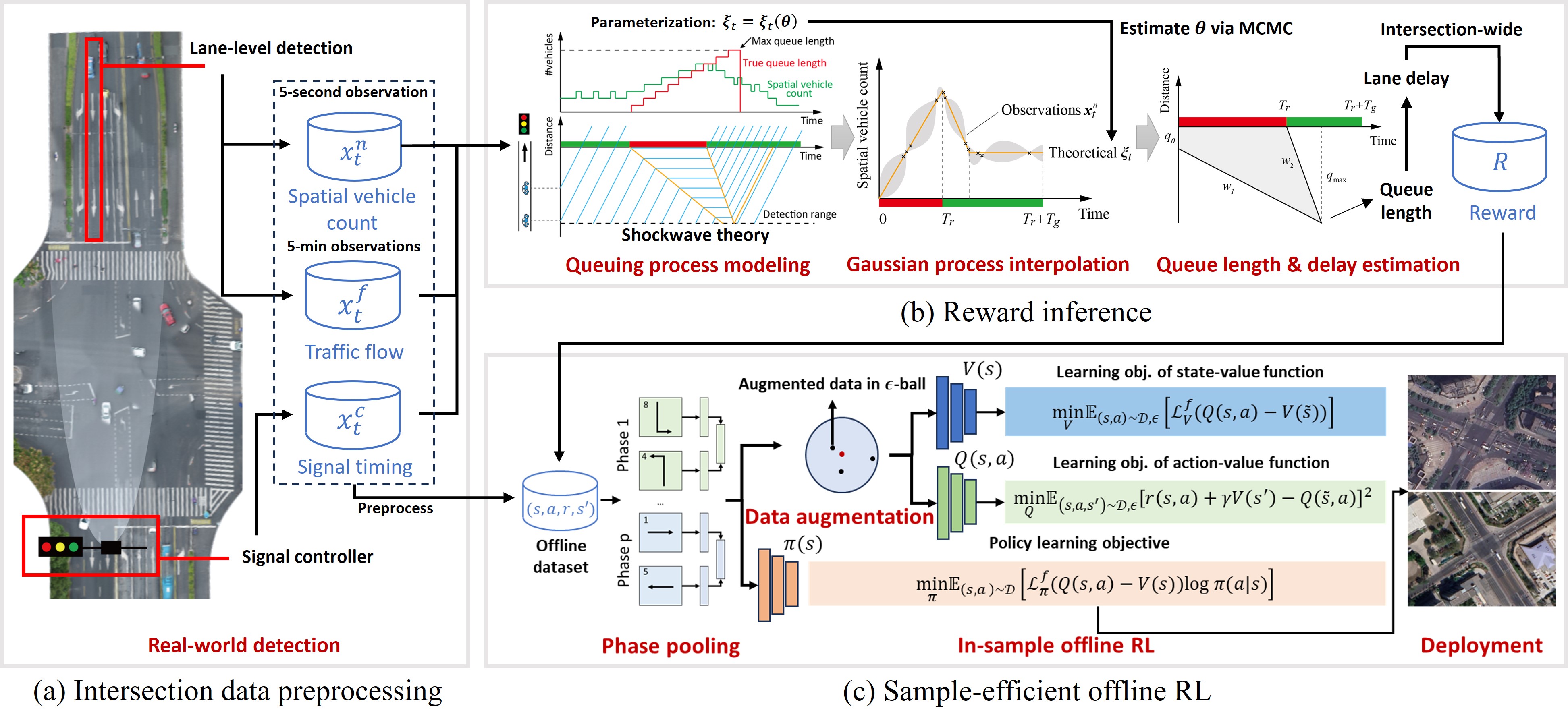}
    \vspace{-8pt}
    \caption{ The proposed D2TSC framework: a fully data-driven approach for realistic traffic signal control using offline RL.}
    \label{fig:Framework}
\end{figure*}

In modern signalized intersections, traffic condition data are usually collected by video cameras mounted on top of traffic light arms. As shown in Figure \ref{fig:Framework}(a), the camera monitors a limited range (typically about 150m) behind the stopline of each entrance lane. It records the spatial vehicle count $\mathbf{x}^n$ within the coverage area and counts the total flow $\mathbf{x}^f$ at some fixed time intervals (in our real-world data, $\mathbf{x}^n$ and $\mathbf{x}^f$ are recorded in 5-minute intervals).
Additionally, signal timing plan $\mathbf{x}^c$ of each signal cycle
is also obtained from the signal controller. 
As can be noted, the real-world obtainable traffic detection data are very coarse-grained. Informative states (e.g., fine-grained queue lengths at each lane) and important performance measures (e.g., intersection delay) for signal control optimization are not directly available, despite that such information is easily accessible in traffic simulators.





\section{Methodology} \label{sec:3}

To develop a deployable RL-based solution for real-world traffic signal control, one needs to address two core technical challenges: 1) optimizing a TSC policy without directly available reward information; and 2) learning with coarse-grained and potentially limited offline data. In this paper, we develop the D2TSC framework to tackle the above challenges, which is illustrated in Figure~\ref{fig:Framework}. D2TSC leverages real-world intersection data for TSC and consists of two key modules: 1) a reward inference model that extracts queue lengths and delay-based rewards from the coarse-grained TSC data by combining traffic flow theory and machine learning; and 2) a sample-efficient offline RL method that enables stable policy learning from coarse-grained and limited real-world data.

\subsection{State and action designs}

\noindent\textbf{1) States}: We sample both the traffic flows $\mathbf{x}^f_t=\left[x^{f}_{t,1},\cdots, x^{f}_{t, L}\right]\in \mathbb{R}^{1\times L}$ and spatial vehicle counts $\mathbf{x}^n_t=\left[x^{n}_{t,1},\cdots,x^{n}_{t,L}\right]\in \mathbb{R}^{1\times L}$ for each lane in 5-min intervals as a part of the states in our RL problem, where $L$ is the set of all lanes in the intersection.
While these features provide coarse-grained information, they are considerably easier to obtain from real-world traffic sensors as compared to information-rich state observations that are only obtainable from traffic simulators. Overall, the raw state features include:
\begin{equation}
s_t=[\mathbf{x}^f_t, \mathbf{x}^n_t]\in\mathbb{R}^{1\times 2L}.
\end{equation}

\noindent\textbf{2) Actions}: For general $|\mathcal{P}|$-phase intersections, we record the cycle length $T_c\in\mathbb{R}$, and the green time ratio for each phase of the timing plan $T_\mathbf{g}=[T_{g}^1, T_g^2, ..., T_g^{|\mathcal{P}|-1}, 1-\sum_{p=1}^{|\mathcal{P}|-1}T_g^p]\in\mathbb{R}^{1\times |\mathcal{P}|}$, where $T_{g}^p$ denotes the green time ratio of the $p$-th phase of the timing plan. Therefore, the green time ratios should sum to 1 and be positive. Overall, we can uniquely define a signal timing plan given a specific cycle length $T_c$ and the green time ratio for each phase $T_\mathbf{g}$, and the action features at step $t$ include:

\begin{equation}
    a_t=[T_c, T_\mathbf{g}]\in\mathbb{R}^{1\times(1+|\mathcal{P}|)}.
\end{equation}

\subsection{Reward Inference}
A key difficulty of using data-driven RL for realistic traffic signal control lies in the absence of ground truth reward information. Performance evaluation metrics that are crucial for TSC, such as reduction in queue lengths and intersection delays, are not directly available from real-world observational data.
To address the problem, we propose a novel reward inference model by combining domain knowledge and machine learning. As illustrated in Figure \ref{fig:Framework}(b), we first model the queuing process based on the shockwave theory~\citep{lighthill1955kinematic,richards1956shock} from transportation engineering and derive a parameterized form of theoretical spatial vehicle count. The associated traffic flow parameters can be estimated by fitting a Gaussian process interpolation model to data. With the estimated parameters, we can further infer the actual queue lengths and delays by analyzing the shockwave boundaries.

\begin{figure*}[t]
    \centering
    \includegraphics[width=\textwidth]{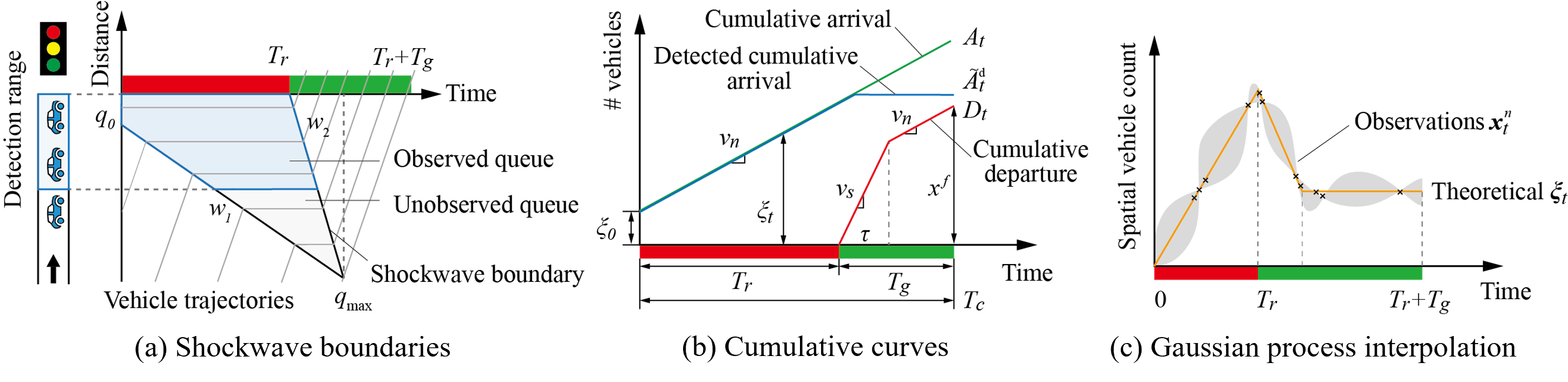}
    \vspace{-8pt}
    \caption{\small{Illustration of queuing process on a lane. (a) Arriving vehicles queue in front of the stopline, forming two shockwaves that propagate backward with speeds $w_1$ and $w_2$. Given an initial queue length $q_0$, the maximum queue length $q_{\text max}$ is formed where the two shockwaves meet. However, in real TSC data, we can only observe the number of vehicles within the detection range, causing the queue possibly be partially observed.
    (b) Cumulative arrival ($A_t$) and departure ($D_t$) curves of vehicles within a signal cycle. Assuming arrival and saturated departure flow rates $v_n$ and $v_s$ remain constant within a signal cycle, based on flow conservation and the impact of shockwaves, the detected cumulative arrival curve $\tilde{A}_t^{\text d}$ lower-bounds the actual cumulative arrival curve. This can be used to construct a theoretical spatial vehicle count curve $\xi_t(\theta)$ based on a set of traffic flow parameters (i.e., $(v_n, v_s, \xi_0)$) and the signal timing information ($T_c, T_g, T_r$). (c) A Gaussian process interpolation model can be constructed to learn $\theta$ by fitting $\xi_t(\theta)$ to the empirically observed spatial vehicle counts $x^n_t$ and traffic flows $x_t^f$.}}
    \label{fig:QueuingProcess}
\end{figure*}


\subsubsection{Queuing process modeling} \label{sec:4.2.1}
The lane queuing process in a signal cycle is illustrated in Figure \ref{fig:QueuingProcess}. Assuming a constant vehicle arrival rate, the cumulative vehicle arrival $A_t$ during the cycle can be characterized by the arrival flow rate $v_n$ and the initial spatial vehicle count $\xi_0$. The cumulative vehicle departure $D_t$ describes the total number of vehicles leaving the lane. As depicted in Figure \ref{fig:QueuingProcess}(b), during the red time, no vehicles leave the lane, whereas during the green time, the process of vehicle departure can be divided into two stages \citep{zhan2020link}. In the first stage, the queued vehicles are released, constituting the saturated flow with a flow rate of $v_s > v_n$. This stage occurs from the beginning of green time until all queued vehicles are fully dissipated ($T_r < t \leq T_r + \tau$). Once all queued vehicles have left ($T_r + \tau < t \leq T_c$), the departure rate becomes the same as the arrival rate $v_n$. Thus, the arrival and departure processes are expressed as: 
\begin{equation}
    \label{equ:CumArrival}
    \begin{aligned}
    &A_t = \xi_0 + v_n t \quad \quad \quad \quad \quad \quad \quad \ \ 0 \leq t \leq T_c \quad \quad \quad \ \\
    \end{aligned}
    \end{equation}
\begin{equation}
    \label{equ:CumDepart}
    \begin{aligned}
    &D_t = \begin{cases}
    0 & 0 \leq t \leq T_r \\
    v_s (t - T_r) & T_r < t \leq T_r + \tau \\
    v_s \tau + v_n (t - T_r - \tau) & T_r + \tau < t \leq T_c \\
    x^f & t = T_c
    \end{cases}
    \end{aligned}
\end{equation}

Ideally, due to flow conservation, the number of vehicles in a lane at time step $t$ is equal to the difference between the cumulative arrival and departure curves (i.e., $A_t-D_t$). However, in real spatial vehicle count data $x^n_t$, we can only observe vehicles located in the detection range with length $L_{dr}$. To account for this restriction, we model the theoretical spatial vehicle count $\xi_t := \tilde{A}_t^{\text d} - D_t$ by introducing the concept of detected cumulative arrival $\tilde{A}_t^{\text d}$, which refers to the cumulative number of vehicles that have entered the detection range. 

Obviously, as illustrated in Figure \ref{fig:QueuingProcess}(b), we have 
\begin{equation}
    \label{equ:detectedArrival}
    \tilde{A}_t^{\text d}\leq A_t
\end{equation}
as the observed cumulative arrival at each lane does not exceed the actual value. Also, according to the Newell's kinetic theory~\citep{newell1993simplified}, the detected cumulative arrival $\tilde{A}_t^{\text d}$ and the cumulative departure $D_t$ within a detected lane section of a given length have the following relationship:
\begin{equation}
    \label{equ:NewellKinetic}
    \tilde{A}_t^{\text d} \leq D_{t-t^S} + L_{dr} k_j
\end{equation}
where $t^S$ is the shockwave propagation time required for traversing the detection range with a length of $L_{dr}$, i.e., $t^S = L_{dr} / w_2$. $k_j$ is the jam density. They are hyperparameters that reflect the inherent traffic flow characteristics of the laneand can be derived from the fundamental diagram (FD), which will be detailed in Section \ref{sec:4.2.4}. 

Due to the limited detection range, it is common for the lane queue length to exceed the detection range during the red time, resulting in a "local spillback". In the event of a local spillback, newly arriving vehicles cannot enter the detected range until the queue tail begins to dissipate. Consequently, Eq. \eqref{equ:NewellKinetic} specifies that the detected cumulative arrival at any time $t$ should be constrained by its cumulative departure condition of a shockwave propagation time $t^S$ earlier. Combining Eqs. \eqref{equ:detectedArrival} and \eqref{equ:NewellKinetic}, the detected cumulative arrival $\tilde{A}_t^{\text d}$ of each lane can be approximated as
\begin{equation}
    \label{equ:detectedArrivalApprox}
    \tilde{A}_t^{\text d}\approx \min \{A_t, D_{t-t^S} + L_{dr} k_j \}
\end{equation}

By substituting Eqs. \eqref{equ:CumArrival} and \eqref{equ:CumDepart} into Eq. \eqref{equ:detectedArrivalApprox}, we have 

\begin{equation}
    \begin{aligned}
    \label{equ:TheoreticalCount}
    \xi_t :=& \tilde{A}_t^{\text d} - D_t \\
    \approx& \min \{ A_t - D_t, D_{t - t^S} - D_t + L_{dr} k_j \} \\
    =& \min \{ \xi_t^{(1)}, \xi_t^{(2)} + L_{dr} k_j \}
    \end{aligned}
\end{equation}
where

\begin{equation}
    \xi_t^{(1)} = \xi_0 + \begin{cases} \begin{aligned}
        &v_n t, & 0 \leq t \leq T_r \\
        &-v_s T_r + (v_n - v_s) t, & T_r < t \leq T_r + \tau \\
        &v_n T_c - x^f, & T_r + \tau < t \leq T_c \\
        \end{aligned} \end{cases}
\end{equation}
\begin{equation}
\resizebox{\linewidth}{!}{
$\xi_t^{(2)} = \begin{cases} \begin{aligned}
        & 0, & 0 \leq t \leq T_r \\
        & - v_s (t - T_r), & T_r < t \leq T_r + t^S, \ t^S < \tau \\
        & - v_s t^S, & T_r + t^S < t \leq T_r + \tau, \ t^S < \tau \\
        & - v_s (t - T_r), & T_r < t \leq T_r + \tau, \ t^S \geq \tau \\
        & v_n T_c - v_s (t^S + T_r) - x^f + (v_s - v_n) t,  
        & T_r + \tau < t \leq T_r + \tau + t^S, \ t^S < T_g - \tau \\
        & - v_n t^S, & T_r + \tau + t^S < t \leq T_c, \ t^S < T_g - \tau \\
        & - x^f + v_n T_c - v_n t, & T_r + \tau < t \leq T_r + t^S, \ t^S \geq T_g - \tau \\
        & - v_s (t^S + T_r) + v_n T_c - x^f + (v_s - v_n) t & T_r + t^S < t \leq T_c, \ t^S \geq T_g - \tau \\
        \end{aligned} \end{cases}$}
\end{equation}
and 
\begin{equation}
\label{equ:Tau}
    \tau = \frac{x^f - v_nT_g}{v_s - v_n} \in [0, T_g]
\end{equation}

Eq. \eqref{equ:TheoreticalCount} introduces a parameterized model for the lane queuing process in a signal cycle with a partial detection range. This model facilitates the comprehensive representation of the theoretical spatial vehicle count $\xi_t$, as a function of a series of traffic flow parameters $\mathbf{\theta} = \{v_n, v_s, \xi_0\}$, i.e., $\xi_t = \xi_t(\mathbf{\theta})$.

\subsubsection{Extension to multi-cycle queuing process}\label{sec:4.2.3}
In order to derive the parameterized queuing process $\xi_t(\mathbf{\theta})$, it is necessary to have access to the cycle-level traffic flow $x^f$ of each lane. However, in practice, the lane-based traffic flow is often collected at fixed time intervals (such as 5 minutes), thereby making it challenging to directly acquire traffic flow for every signal cycle. To address the issue, we provide a simplistic estimation approach to extend the queuing process modeling to multi-cycle situations. 

\begin{figure*}[t]
\begin{minipage}{.55\textwidth}
    \includegraphics[width=\linewidth]{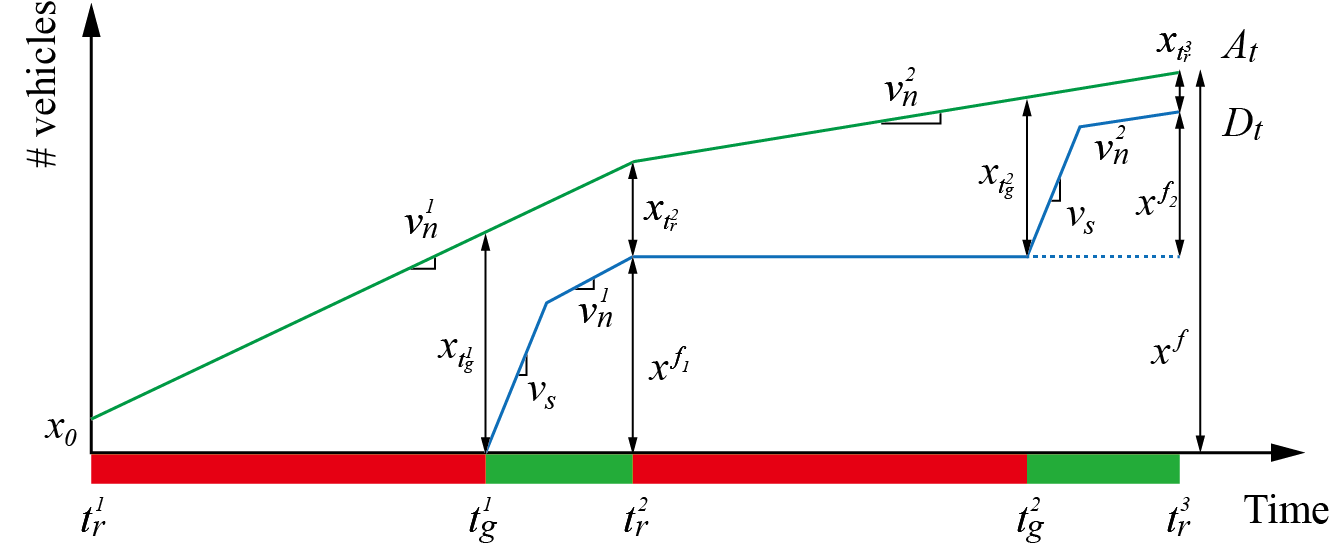}
    \caption{General queuing process in multiple signal cycles.}
    \label{fig:multi-cycleQueuing}
\end{minipage}
\quad\quad
\begin{minipage}{.345\textwidth}
    \includegraphics[width=\linewidth]{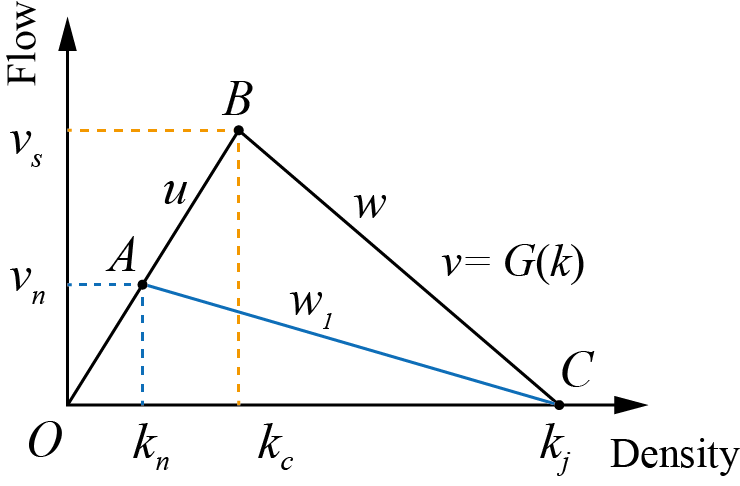}
    \caption{Fundamental diagram of traffic flow.}
    \label{fig:FD}
\end{minipage}
\end{figure*}

To estimate $\mathbf{\xi}_t(\mathbf{\theta})$ using lane-based traffic flow  across multiple signal cycles,
we maintain the assumption of constant arrival rate within each signal cycle, additionally, we assume that the arrival rate of each signal cycle is proportional to the maximum observed spatial vehicle count in the cycle. Figure \ref{fig:multi-cycleQueuing} illustrates the general queuing process for multi-cycle situations. Suppose that there are $C$ complete signal cycles during the flow detection interval, and, with a slight abuse of notations, let $x^f$ represent the total flow in these cycles. We denote the arrival rate, cycle flow, cycle beginning time, red and green durations, and cycle length of the $c$-th cycle as $v_n^c$, $x^{f,c}$, $t_r^c$, $T_r^c$, $T_g^c$, and $T_c^c$, respectively. Based on the proportionality assumption of cycle arrival rates, we have the following relationship: 
\begin{equation}
    \label{equ:MSQCons}
    \begin{aligned}
    \frac{v_n^c}{\max_{t_r^c\leq t < t_r^{c+1}}(x^n_t)} &= \frac{v_n^{c^{\prime}}}{\max_{t_r^{c^{\prime}}\leq t < t_r^{c^{\prime}+1}}(x^n_t)} \\& =  \zeta, 
    \quad \forall c, c^{\prime} \in \{1, \cdots, C\}
    \end{aligned}
\end{equation}
where $\zeta$ is the normalized arrival rate. According to Figure \ref{fig:multi-cycleQueuing}, the cycle and total traffic flows should satisfy the following conservation conditions, 
\begin{equation}
    \label{equ:TotalFlowCons}
    x^f = \sum_{i=1}^C x^{f,c} + x^{n_r}
\end{equation}
\begin{equation}
    \label{equ:CycleFlowCons}
    x^{f,c} = x^n_{t_r^c} + v_n^c T_c^c - x^n_{t_r^{c+1}}, \quad \forall c \in \{1, \cdots, C\}
\end{equation}
where $x^{n_r}$ is the remaining spatial vehicle count at the end of the flow detection interval. Equations \eqref{equ:MSQCons}-\eqref{equ:CycleFlowCons} yield a solution $\{x^{f,c}\}$. Consequently, the general multi-cycle estimation problem can be simplified into multiple independent single-cycle queuing process problems. One can estimate the parameters $\mathbf{\theta}$ for each lane in each signal cycle by fitting it to the empirical spatial vehicle counts $x_t^n$. To this end, we develop a non-parametric approach to find the most likely theoretical spatial queue count curve using a Gaussian process interpolation model in the next section.

\subsubsection{Gaussian process interpolation}
The Gaussian process interpolation technique is a favorable tool within the realm of Bayesian statistical modeling and machine learning. It establishes a probability distribution over functions, linking output $y$ to input $\mathbf{x}$, wherein the values of $y$ for any set of inputs ${\mathbf{x}_1, \mathbf{x}_2, \cdots, \mathbf{x}_N}$ jointly conform to a Gaussian distribution. As illustrated in Figure \ref{fig:QueuingProcess}(c), we can model the empirical observations $x_t^n$ as the parameterized theoretical spatial vehicle count curve $\xi_t(\theta)$ with Gaussian disturbances, i.e.,
\begin{equation}
    \mathbf{x}_\mathbf{t}^n(\mathbf{\theta}) = \mathbf{\xi}_\mathbf{t}(\mathbf{\theta}) + \eta \mathbf{\varepsilon},\label{eq:GP_mean}
\end{equation}
where $\mathbf{\varepsilon} \sim N(0, \mathbf{I})$ represents a Gaussian distributed disturbance, and $\eta$ is a scale hyperparameter. Then it will lead to a Gaussian process:  
\begin{equation}
    GP(\mathbf{t} | \mathbf{\theta})\sim N(\mathbf{\xi}_\mathbf{t}(\mathbf{\theta}), \textit{Ker}(\mathbf{t, t | \eta}))
\end{equation}
which is parameterized by the mean process of $\mathbf{\xi}_\mathbf{t}(\mathbf{\theta})$ and a covariance kernel matrix $\textit{Ker}(\mathbf{t, t | \eta}))$, defined as
\begin{equation}
    \textit{Ker}(\mathbf{t, t | \eta}) = 
    \begin{bmatrix}
    \phi(t_1, t_1|\eta) & \dots & \phi(t_1, t_n|\eta) \\
    \vdots & \ddots & \vdots \\
    \phi(t_n, t_1|\eta) & \dots & \phi(t_n, t_n|\eta) 
    \end{bmatrix} \\
\end{equation}
where $\phi(\cdot|\eta)$ is the kernel function used to measure the covariance between time stamps of a pair of observed spatial vehicle count $x^n_{t_i}$ and  $x^n_{t_j}$ in a signal cycle. We adopt the squared exponential function with the regulation term as the kernel function: 
\begin{equation}
\begin{split}
    \phi(t_i, t_j|\eta) = h_0 e^{-{\Big(\frac{t_i - t_j}{\lambda}\Big)}^2} + \eta^2 \delta(t_i, t_j) 
\end{split}
\end{equation}
where $\delta(t_i, t_j) = 1$ if $t_i = t_j$ and 0 otherwise, and $h_0$, $\lambda$ are scale hyperparameters. Essentially, $\mathbf{x}_\mathbf{t}^n(\mathbf{\theta})$ in Eq. (\ref{eq:GP_mean}) can be regarded as the observed spatial vehicle count $\mathbf{x}_\mathbf{t}^n$ drawn at timestamps $\mathbf{t}$ from a multivariate Gaussian distribution parameterized by $\mathbf{\theta}$ and $\eta$. 
Therefore, the traffic parameters $\mathbf{\theta}$ of a signal cycle that parameterizes the most likely theoretical spatial vehicle count $\xi_{\mathbf{t}}$ given the observations $x^n_{\mathbf{t}}$ can be learned by maximizing the joint probability distribution of $GP(\mathbf{t} | \mathbf{\theta})$. 

Inferring queuing process parameters $\mathbf{\theta}$ requires performing maximum likelihood estimation on  Gaussian process $GP(\mathbf{t} | \mathbf{\theta})$. However, the joint probability distribution $GP(\mathbf{t} | \mathbf{\theta})$ is complex given a non-differentiable piece-wise linear function $\mathbf{\xi}_t(\mathbf{\theta})$ in Eq. \eqref{equ:TheoreticalCount}. To this end, we develop a Metropolis-Hastings (M-H) algorithm that efficiently infers $\mathbf{\theta}$ using the using Markov Chain Monte-Carlo (MCMC) technique. M-H algorithm is a widely used sampling method, which is useful in estimating parameters from complex probability distributions~\citep{bishop2006pattern, berger2013statistical}. The developed M-H algorithm for learning parameters $\theta$ is given in Table \ref{tab:MH}. 

\begin{table}[h]
    \caption{M-H algorithm for $\theta$ inference.}
    \centering
    \label{tab:MH}
    \renewcommand{\arraystretch}{1.2}
    \setlength{\tabcolsep}{3mm}{}
    \begin{tabularx}{\linewidth}
        {>{\raggedright\arraybackslash}X}
        \hline
        \textbf{Input}: Timestamps $\mathbf{t}$, the observed spatial vehicle count $\mathbf{x}^n$, and the cycle flow $x^f$. \\
        \textbf{Output}: Estimated parameters $\theta$ \\
        \hline
        \textbf{Step 1}: Initialize $\theta^{(0)}$ \\
        \textbf{Step 2}: Sample $\hat \theta$ from a uniform distribution \\
        \textbf{Step 3}: Computing the likelihood of $\hat \theta$ and accept it as $\theta^{(k)}$ according to the probability of 
        $\beta = \min(p(\mathbf{x}^n|\mathbf{\hat\theta}) / p(\mathbf{x}^n |\mathbf{\theta}^{(k-1)}), 1)$ \\
        \textbf{Step 4}: Repeat steps 2-3 for a given number of iterations until $\theta^{(k)}$ is stable \\
        \textbf{Step 5}: Discard the first 75\% of accepted samples as burn-in, and use the mean of remaining values as the result of $\theta$ \\
        \hline
    \end{tabularx}
\end{table}

\subsubsection{Queue length and delay estimation} \label{sec:4.2.4}

Once we have estimated the queuing process parameters $\theta=\{v_n, v_s, \xi_0\}$, we can further use them to infer the lane-level queue length and delay in each signal cycle. According to the geometric relationship illustrated in Figure \ref{fig:QueuingProcess}(a), the maximum queue length $q_{\text{max}}$ of each lane can be estimated by 
\begin{equation}
    \label{equ:MaxQueue}
    q_{\text{max}} = \frac{w_{2} (q_{0} + w_{1} T_r)}{w_{2} - w_{1}}
\end{equation}
where $w_1$, $w_2$ represent the stopping and starting shockwave speed, respectively, and $q_0 = \xi_0 k_j$ is the initial queue length at the beginning of the cycle, $k_j$ is the jam density of the lane. In the shockwave theory, these parameters can be inferred from the fundamental diagram (FD), which is a common representation of inherent relationship between traffic flow $v$ and density $k$. In this study, we assume that each lane independently satisfies the following flow-density relationship~\citep{daganzo1997fundamentals, jin2015point}, as illustrated in Figure \ref{fig:FD}: 
\begin{equation}
    \label{equ:TriFD}
    v = G(k) = \min \{uk, w(k_j-k)\}
\end{equation}




The FD hyperparameters include the jam density $k_j > 0$, the free-flow speed $u > 0$, and the shockwave speed $w > 0$. According to the shockwave theory, the stopping shockwave speed in Figure \ref{fig:QueuingProcess}(a) during the red time can be calculated from the slope of the secant line AC on the FD curve, 
\begin{equation}
\begin{aligned}
    w_1 = k_{AC} &= \frac{v_n}{k_n - k_j} = \frac{1}{1 / u - k_j / v_n} \\&= - \frac{1}{k_j (1 / v_n - 1 / v_s) + t^S / L_{dr}}
\end{aligned}
\end{equation}
Here, $w_1 < 0$ suggests that the shockwave propagates backward (opposite to the driving direction of vehicles). Similarly, the starting shockwave speed can be obtained by
\begin{equation}
    w_2 = k_{BC} = -w
\end{equation}

With $w_1$, $w_2$ and $q_0$ being parameterized by the estimated queue length parameters $\theta=\{v_n, v_s, \xi_0\}$ and static FD hyperparameters $\{k_j, u, w\}$, we can conveniently calculate maximum queue length $q_{\textit{max}}$ of each lane in each signal cycle using Eq. (\ref{equ:MaxQueue}).
The cycle-level total delay $d$ of each lane (total waiting time spent in the queue) can thus be estimated as the area enclosed by the shockwave boundaries as shown in Figure \ref{fig:QueuingProcess}(a), mathematically, 
\begin{equation}
    \label{equ:LaneTotalDelay}
    d = \frac{1}{2} \Big[ T_r q_{\text{max}} + \frac{q_{0} (w_{2} T_r + q_{0}) }{w_{2} - w_{1}} \Big]
\end{equation}




\subsubsection{Reward function parameterization}

To be compatible with coarse-grained traffic flow data, we design our RL agent to make decisions in 5-min intervals. Suppose that there are $C_t$ complete signal cycles within the $t$-th 5-min interval,
we define the reward as the average vehicle delay within these cycles. In real-world traffic signal control systems, average vehicle delay is commonly used as a direct indicator of the level of service (LoS) of an intersection. We can thus calculate the rewards by weighting the cycle total flow $x_{t, l}^{f,c}$ to the cycle total delay $d_{l, c}$ of lane $l$ of signal cycle $c$, namely, 
\begin{equation}
    \label{equ:reward}
    r_t = \frac{ \sum_{l=1}^{L} \sum_{c=1}^{C_t} d_{l, c} x_{t, l}^{f,c} }{ \sum_{l=1}^{L} \sum_{c=1}^{C_t} x_{t, l}^{f,c} }
\end{equation}
Both $x_{t,l}^{f,c}$ and $d_{l, c}$ in Eq. \eqref{equ:reward} can be estimated from the coarse-grained information using the proposed techniques in Sections \ref{sec:4.2.3} and \ref{sec:4.2.4}. Finally, these reward values will be normalized and used for offline RL policy training.


\subsection{Signal Control Optimization via Offline RL}
With the inferred rewards, we now present the proposed sample-efficient offline RL method used in D2TSC. The coarse-grained and potentially limited real-world TSC data poses a great technical challenge for offline policy learning.
To tackle this, we first introduce a phase pooling design to facilitate effective traffic demand information extraction under the time-varied phase orders
(Section~\ref{sec:sample_efficient}). Then, we employ the SOTA in-sample learning offline RL algorithm to achieve stable and high-performing policy learning (Section~\ref{sec:in_sample}). In addition, we propose a data augmentation scheme (Section~\ref{sec:data_aug}) to improve the sample efficiency and the OOD-generalization ability of the learned TSC policy given limited available data.

\subsubsection{Phase information modeling}\label{sec:sample_efficient}
For intersections that have $|\mathcal{P}|$ phases with $K$ varied phase orders, it is challenging to accurately grasp the traffic demand for each phase solely from the limited coarse-grained data, as the same lane may belong to different phases in different time periods. This challenge imposes stringent requirements on data quantities and qualities for RL agents to learn good TSC policies. However, in practice, the available offline data are coarse-grained and typically have narrow state-action space coverage,
which makes it hard to distill informative features for RL policy learning. Therefore, it is crucial to incorporate some domain knowledge to facilitate RL policy learning.

In this work, we design a \textit{phase pooling layer} $\textit{PPL}(\cdot)$ to facilitate the RL agent to better understand the intersection-wide traffic pattern. 
Specifically, for a specific phase order, we aggregate the traffic information of each phase from the coarse-grained features. 
In details, all $K$ phase orders could be modeled in a metric $\Phi_\mathbf{M}\in\mathbb{R}^{K\times |\mathcal{P}| \times L}$ that contains only binary values, where $\Phi_\mathbf{M}(k, p, l)=1$ represents that the $l$-th lane should display green at the $p$-th phase for the $k$-th phase order and $\Phi_\mathbf{M}(k, p, l)=0$ denotes red signal. In practice, the intersection will select a phase order $k$ and follow the orders defined in $\Phi_\mathbf{M}$ to display green or red signals for each lane for a period of time and then switch to another phase order according to some human-defined rules. To model this, we also incorporate a one-hot embedding of the phase order ID $\Phi_{\rm ID}\in\mathbb{R}^{1\times K}$ to indicate the specific phase order that the intersection currently undergoes. For instance, $\Phi_{\rm ID}(k)=1$ means the intersection currently undergoes the $k$-th phase order. 
In this case, $\Phi_{\rm ID}  \Phi_\mathbf{M}\in\mathbb{R}^{|\mathcal{P}|\times L}$ can uniquely determine a phase pattern. Then, $\Phi_{\rm ID}  \Phi_\mathbf{M}(p, l)=1$ means that the current intersection has green signal at the $p$-th phase for the $l$-th lane and $\Phi_{\rm ID}  \Phi_\mathbf{M}(p, l)=0$ corresponds a red signal. 
The phase pooling layer is thus defined as:
\begin{equation}
    \textit{PPL}(\mathbf{x}^f_t, \Phi_{\rm ID}, \Phi_\mathbf{M})=\mathbf{x}^f_t (\Phi_{\rm ID}\Phi_\mathbf{M})^T\in\mathbb{R}^{1\times |\mathcal{P}|},
\end{equation}
where the $p$-th item of $\mathbf{x}^f_t (\Phi_{\rm ID}\Phi_\mathbf{M})^T$ represents the aggregated flow traffic demand information of the $p$-th phase (sum of the flow information for all the lanes that display green signal at the $p$-th phase). This provides more direct phase-based information to facilitate TSC RL policy learning, 
as a high traffic demand for a phase often suggests a larger green time ratio should be allocated for this phase. Finally, the augmented state features at step $t$ include:
\begin{equation}
    \hat{s}_t=\left[\mathbf{x}^f_t, \mathbf{x}^n_t, \Phi_{\rm ID}, \textit{PPL}(\mathbf{x}^f_t, \Phi_{\rm ID}, \Phi_\mathbf{M})\right]\in\mathbb{R}^{1\times(2L+K+|\mathcal{P}|)}
\end{equation}

\subsubsection{In-sample offline RL algorithm}\label{sec:in_sample}

In this paper, we consider the behavior regularized MDP framework in Eq.~(\ref{equ:brMDP}) for offline TSC policy learning and turn the optimization objectives in Eq.~(\ref{equ:ac_q}-\ref{equ:ac_pi}) to its in-sample learning form for its great training stability and SOTA performances. As formally studied in \citet{sql}, the behavior regularized MDP is closely related to a class of state-of-the-art (SOTA) \textit{in-sample learning} offline RL methods~\citep{iql, sql, garg2023extreme,hansen2023idql}.  Specifically, Eq.~(\ref{equ:ac_q}-\ref{equ:ac_pi}) can be easily transferred to its in-sample version by solving the KKT condition of Eq.~(\ref{equ:ac_q}-\ref{equ:ac_pi})~\citep{sql}.
In detail, different choices of the $f$-function in Eq.~(\ref{equ:brMDP}) will lead to different \textit{in-sample learning} algorithms, but they all share the following general learning objectives:
\begin{align}   &\min_V\mathbb{E}_{(s,a)\sim\mathcal{D}}\left[\mathcal{L}_{V}^f\left(Q(s,a)-V(s)\right)\right]\label{equ:train_V_main},\\
    &\min_{Q}\mathbb{E}_{(s,a,s')\sim\mathcal{D}}\left[r(s,a)+\gamma V(s')-Q(s,a)\right]^2\label{equ:train_Q_main},\\
    &\min_{\pi}\mathbb{E}_{(s,a)\sim\mathcal{D}}\left[\mathcal{L}_{\pi}^f(Q(s,a)-V(s))\log\pi(a|s)\right],\label{equ:train_pi_main}
\end{align}
where $Q(s,a)$ and $V(s)$ are the action-value function and state-value function, respectively;
the forms of $\mathcal{L}_V^f(\cdot)$ and $\mathcal{L}_{\pi}^f(\cdot)$ depend on the specific choice of $f$-function. For instance, we can choose $f(x)=\log(x)$, where $\mathcal{L}_{V}^f(x)=\exp\left(x/\alpha\right)-x/\alpha$ and $\mathcal{L}_{\pi}^f(x)=\exp\left(x/\alpha\right)$. This constructs a standard KL-divergence regularized MDP~\citep{haarnoja2018soft} and is equivalent to an in-sample learning method, Exponential Q-Learning (EQL)~\citep{sql, garg2023extreme, hansen2023idql}. However, the exponential term in EQL typically faces numerical instability, which requires additional regularization tricks to stabilize the training. Therefore, we choose $f(x)=x-1$ in this paper instead, which is equivalent to another SOTA in-sample learning offline RL method, Sparse Q-Learning (SQL)~\citep{sql}, and solves a Neyman $\chi^2$-divergence regularized objective~\citep{kumar2020conservative}, from which $\mathcal{L}_V^f(\cdot)$ and $\mathcal{L}_{\pi}^f(\cdot)$ in the general learning objectives in Eq.~(\ref{equ:train_V_main})-(\ref{equ:train_pi_main}) become:
\begin{align}
    &\mathcal{L}_{V}^f(x)=\mathbb{I}(1+x/2\alpha>0)(1+x/2\alpha)^2-x/\alpha,\label{equ:lv}\\
    &\mathcal{L}_{\pi}^f(x)=\mathbb{I}(1+x/2\alpha>0)(1+x/2\alpha),\label{equ:lpi}
\end{align}
where $\mathbb{I}(\cdot)$ is the indicator function. Observe in Eq.~(\ref{equ:lv}) that the state-value function ($V(s)$) learning objective seeks to regress only on high Q-values $Q(s,a)$ where $1+x/2\alpha>0$, which can implicitly find the optimal state-value function covered by the offline dataset~\citep{sql}. Meanwhile, Eq.~(\ref{equ:lpi}) shows that the policy behaves in a weighted regression manner~\citep{nair2020awac} that only maximizes the likelihood on the regions that attains high values, thus producing optimized policy that leads to good outcomes. 

Optimizing the in-sample learning objectives in Eq.~(\ref{equ:train_V_main}-\ref{equ:train_pi_main}) enjoys several advantages compared to directly optimizing Eq.~(\ref{equ:ac_q}-\ref{equ:ac_pi}). First, note that minimizing the objectives in Eq.~(\ref{equ:train_V_main}-\ref{equ:train_pi_main}) does not need to explicitly learn a behavior policy $\mu$ to enforce policy regularization with respect to offline data, thus bypassing the notoriously difficult behavior estimation problem~\citep{nair2020awac}. 
Second, optimizing the in-sample learning objectives in Eq.~(\ref{equ:train_V_main}-\ref{equ:train_pi_main}) involves only samples from the offline dataset $\mathcal{D}$, without any counterfactual reasoning on potential OOD policy-induced samples, thereby offering
great learning stability as compared to other types of offline RL methods. On the contrary, the objectives in Eq.~(\ref{equ:ac_q}-\ref{equ:ac_pi}) are susceptible to overestimation errors at OOD regions. Specifically, Eq.~(\ref{equ:ac_q}) involves inferring action value function $Q(s',a')$ for actions $a'$ generated by the policy $\pi$, which may not present in the offline dataset and can easily lead to erroneous value estimation.
Considering these advantages, we choose to turn Eq.~(\ref{equ:ac_q}-\ref{equ:ac_pi}) to its corresponding \textit{in-sample learning} offline RL methods to solve the \textit{behavior regularized} MDP instead in this paper. This facilitates learning the optimal policy using only data that is seen in the offline dataset, avoiding the difficult behavior estimation and bypassing the potential over-estimation issue that arises from the counterfactual reasoning at OOD regions.

\subsubsection{Improving sample-efficient via data augmentation}
\label{sec:data_aug}
By optimizing Eq.~(\ref{equ:train_V_main}-\ref{equ:train_pi_main}), we can in principle acquire a good TSC policy given enough offline data with good coverage over the state-action space. However, \textit{in-sample learning} may exhibit poor generalization capability in OOD regions during deployment due to the absence of supervision signal on OOD regions~\citep{li2023when}. Furthermore, real-world TSC datasets are typically small and have narrow coverage of the state-action space, which further exacerbates the challenge of OOD generalization. 

To address this, we introduce a data augmentation scheme inspired by S4RL~\citep{sinha2022s4rl} that generates augmented data to mitigate the small dataset learning problem and enhance generalization in OOD regions. Specifically, during training of $V$ and $Q$, small Gaussian noises $\epsilon$ are added to the states in the data batch (i.e., $\tilde{s}:=s+\epsilon$), leading to the following augmented training objective:
\begin{align}
    &\min_V\mathbb{E}_{(s,a)\sim\mathcal{D}, \epsilon\sim\mathcal{N}(0,\sigma^2)}\left[\mathcal{L}_{V}^f\left(Q(s,a)-V(\tilde{s})\right)\right],\notag\\
    &\min_{Q}\mathbb{E}_{(s,a,s')\sim\mathcal{D}, \epsilon\sim\mathcal{N}(0, \sigma^2)}\left[\left(r(s,a)+\gamma V(s')-Q(\tilde{s},{a})\right)^2\right]\notag
\end{align}

This data augmentation can be perceived as smoothing the value function in its local $\epsilon$-ball, which is beneficial to combat small perturbations and thus improves generalization on OOD regions. In addition, the augmented data can increase the dataset quantity and thus improve the overall sample efficiency of the offline RL algorithm~\citep{sinha2022s4rl}.




\section{Experiments} \label{sec:4}
\subsection{Experiment Settings}
\subsubsection{Offline datasets and evaluation protocol.}
In this section, we evaluate our proposed D2TSC framework against several 
competitive baselines. Specifically, we collect 7 days (June 16-22, 2023) of historical traffic data from one test intersection in Zhuzhou, China, which is a complex 4-phase intersection with 17 lanes, as shown in Figure~\ref{fig:google_map}. We develop a highly customized simulation environment based on SUMO~\citep{lopez2018microscopic} that strictly follows the real-world traffic characteristics observed in the collected 7 days of historical traffic data to provide comparable evaluations of all methods since it is infeasible to deploy some inferior baselines in real-world intersections for testing, which can be risky. To obtain the offline dataset for policy learning, we generate 100 days of data from the highly customized simulator using the behavior cloning policy trained on the 7 days' real datasets, as collecting a large amount of real-world data is quite costly. Note that the simulator is used only for generating the offline datasets and evaluating the policies, but is not accessible during the training process. Also, the generated data are coarse-grained to align with the realistic TSC data.

\subsubsection{D2TSC hyperparamters}
For the hyperparameters of D2TSC in our experiments, we use 2-layer MLPs to model the $Q$ function, $V$ function, and policy $\pi$ in Eq.~(\ref{equ:train_V_main})-(\ref{equ:train_pi_main}).
The learning rates for all networks are $3e-5$ since the offline datasets are small, from which a large learning rate may lead to severe overfitting. We use the Adam optimizer to optimize the network parameters. We normalize all the rewards and state features in the offline dataset to $\mathcal{N}(5,1)$ and $\mathcal{N}(0,1)$; the actions are normalized to the range of $[0,1]$.
The regularization weight $\alpha$ in Eq.~(\ref{equ:brMDP}) is set to $0.01$, since we observe that a large conservative strength may be over-conservative and lead to unsatisfactory performances. For the {data augmentation scale}, the standard deviation of the added Gaussian noise is 0.01. We also clip the sampled noise to be smaller than 0.025 akin to TD3 and TD3+BC~\citep{fujimoto2018addressing, fujimoto2021minimalist} to avoid unrealistic augmented data caused by the overly large added noise.

\subsection{Baseline Methods}
In this paper, we consider the following baselines for comparison. For all baselines, we keep the network architecture, learning rates, and optimizer the same as D2TSC for a fair comparison. Furthermore, same as D2TSC, we use our inferred rewards to train all RL baselines, as there is no reward signal from the offline TSC dataset, although these methods themselves do not have the capability to learn from real-world reward-free TSC data.
\begin{itemize}[leftmargin=*,topsep=0pt,noitemsep]
  \item  \textbf{Fixed plan} is the fixed background timing plan used in the actual real-world test intersection.
  \item  \textbf{Conventional TSC} refers to a saturation-balance adaptive control method devised by domain experts and has been deployed in the test intersection for commercial usage. Therefore, this approach is quite high-performing and can be regarded as a near-expert policy.
  \item \textbf{Behavior cloning (BC)}~\citep{pomerleau1988alvinn} is a widely used imitation learning method that directly mimics the offline data to learn the policy. 
  \item \textbf{TD3+BC}~\citep{fujimoto2021minimalist} is a well-known policy constraint offline RL method. We re-implement the official codes to be compatible with our simulation evaluation environment.
  We carefully tune its hyperparameters to ensure the best performance.
  \item \textbf{IQL}~\citep{iql} is another SOTA in-sample learning offline RL method that utilizes expectile regression to optimize the value functions and extracts policy using advantage-weighted regression. We also re-implement the official codes and fine-tune its hyperparameters to be suitable to our TSC problem.
  \item \textbf{DataLight}~\citep{zhang2023data} is an offline TSC optimization method built upon the offline RL algorithm CQL~\citep{kumar2020conservative}. We re-implement the official codes and use the original training parameters. We adapt its input states to be the same as our state features. We also modify their action space to be able to control the phase in our setting. 
  \item \textbf{DemoLight (offline)}~\citep{xiong2019learning} is a recent TSC optimization method that utilizes BC pretrain to accelerate online RL training using A2C algorithm~\citep{mnih2016asynchronous}. To be compatible with our offline setting, we consider an offline version of DemoLight that initializes the policy with expert demonstrations using BC, and then continues to run A2C on the fixed offline dataset without online interactions.
\end{itemize}

\subsection{Reward Inference Evaluation}
\label{subsec:5.2}
The proposed reward inference approach is evaluated on our customized simulation environment configured by real-world traffic data.
Vehicle positions and speeds are recorded from the simulator every second (regardless of the detection range). As a result, it is possible to compute the actual delay and queue length. Individual vehicle delay is defined as the cumulative difference between its speed and the free-flow speed over time. The lane delay is calculated by summing all individual vehicle delays occurring in the lane. The maximum queue length of a lane is determined as the farthest position of stopped vehicles (with speeds below 5.4 km/h) in the lane during a signal cycle, corresponding to the maximum range of the shockwave. 

In our model, the Gaussian process hyperparameters are selected as $h_0 = 0.5$, $\lambda = 2$, and $\eta = 1$. The shockwave hyperparameters are $L_{dr} = 150$ m, $k_j = 7.5$ m/veh, and $t^S = 25$s. The estimation of parameters $\mathbf{\theta}$ using the M-H algorithm involves 1000 iterations per signal cycle for each lane. Only lanes associated with signal-controlled traffic movements (i.e., left-turn and straight, excluding right turn) are estimated for the maximum queue length and account for the total delay of the intersection. 

\begin{figure*}[t]
    \centering
    \includegraphics[width=\linewidth]{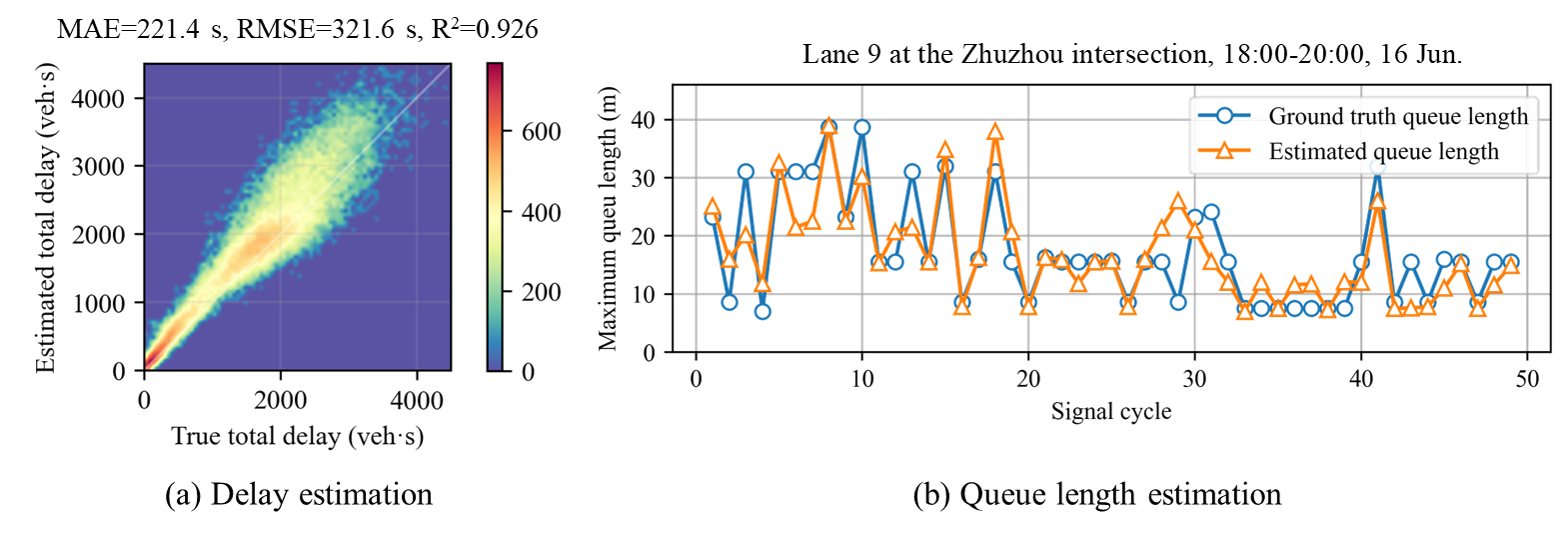}
    \vspace{-14pt}
    \caption{Reward estimation results. (a) 5-minute total delay of the intersection. Colorbars represent the data frequency. (b) A case study of cycle-level lane maximum queue length estimation during peak hours. }
    \label{fig:EstimationResults}
\end{figure*}

Figure \ref{fig:EstimationResults} illustrates the estimation results of the 5-minute total delay for the test intersection. 
The estimated total delay closely aligns with the ground truth delay, exhibiting similar distributions. Note that under worse traffic conditions, certain lanes may experience extreme queuing that exceeds the detection range. This could lead to the underestimation of queuing conditions if directly adopting the observed spatial vehicle counts as queue lengths. In contrast, our proposed estimation approach performs very well at capturing the real delay, mainly due to the establishment of parametric formulations for the actual queuing process. The estimated queue lengths also match well with the ground truth during peak hours. These results demonstrate that our model can provide robust and reliable reward signals for TSC optimization.

\begin{table*}[t!]
    \centering
    \scriptsize
    \caption{\small Optimization results of the test intersection. We report the mean and standard deviation across 5 random seeds.}
    \vspace{-4pt}
    \begin{tabular}{lcccccccccccccccccc}
    \toprule
    \multicolumn{9}{c}{\textbf{Total intersection delay} (the smaller the better)} \\
    \midrule
     \textbf{Method}         &\textbf{16 Jun.} &\textbf{17 Jun.} &\textbf{18 Jun.} &\textbf{19 Jun.} &\textbf{20 Jun.} &\textbf{21 Jun.} &\textbf{22 Jun.} & \textbf{Mean} \\
    \midrule
    Fixed plan & 5953	&6266	&6120	&6194	&6396	&6383	&6480	&6256\\
    \midrule
    Conventional TSC &$4583$           &$4646$	      &$4628$	        &$4591$	        &$4618$	     &$4701$	        &$4515$	 &$4612$  \\
    \midrule
    BC              &$4841\pm 27   $      &$4738\pm 28$    &$4777\pm 19$    &$4629\pm 37$      &$4616\pm 42$   &$4131\pm 48$     &$4876\pm 39$    &$4659\pm 47$\\
    TD3+BC          &$4522\pm12$	&$4714\pm38$  &$4588\pm22$	&$4579\pm40$	&$4634\pm37$ &$4633\pm51$	&$4551\pm27$  &$4603\pm5$\\
    IQL             &$4543\pm36$	&$4675\pm42$  &$4552\pm24$	&$4550\pm46$	&$4622\pm29$ &$4616\pm46$	&$4509\pm52$  &$4581\pm9$\\
    \midrule
    DataLight   &$4542\pm 32$      &$4679\pm 39$    &$4558\pm 48$    &$4549\pm45$      &$4628\pm81$   &$4614\pm48$     &$4507\pm38$    &$4582\pm57$\\
    DemoLight (offline)          &$8851\pm 81$      &$10327\pm84$   &$6118\pm88$    &$11073\pm95$     &$6138\pm93$   &$7480\pm92$     &$8872\pm91$    &$8408\pm101$\\
    \midrule
    D2TSC \textit{w/o} PPL&$4533\pm84$ &$4660\pm99$ &$4580\pm131$ &$4540\pm125$   &$4585\pm118$&$4611\pm99$   &$4522\pm121$  &$4659\pm45$\\
    D2TSC \textit{w/o} AG&$4518\pm30$ &$4659\pm45$ &$4565\pm49$  &$4524\pm33$    &$4617\pm57$ &$4604\pm44$   &$4490\pm43$  &$4568\pm12$\\
    D2TSC (Ours) &{$\bf{4513\pm43}$}	&$\bf{4605\pm45}$  &$\bf{4505\pm45}$	&$\bf{4479\pm80}$	&$\bf{4535\pm68}$ &$\bf{4580\pm55}$	&$\bf{4463\pm52}$  &$\bf{4526\pm37}$\\
    \midrule
    \midrule
    \multicolumn{9}{c}{\textbf{Total queue Length} (the smaller the better)} \\
    \midrule
     \textbf{Method}         &\textbf{16 Jun.} &\textbf{17 Jun.} &\textbf{18 Jun.} &\textbf{19 Jun.} &\textbf{20 Jun.} &\textbf{21 Jun.} &\textbf{22 Jun.} & \textbf{Mean} \\
    \midrule
    Fixed plan &5295	&6339	&5994	&6078	&7090	&6696	&6769	&6323\\
    \midrule
    Conventional TSC         &$3558$           &$3931$	      &$3724$	        &$3627$	        &$3844$	     &$3684$	        &$3384$	 &$3679$  \\
    \midrule
    BC              &$3528\pm 58$      &$3805\pm 63$    &$3553\pm 76$    &$3931\pm 54$      &$3808\pm 58$   &$3552\pm 57$     &$3427\pm 32$    &$3658\pm 89 $\\
    TD3+BC          &$3392\pm67$	&$3922\pm76$  &$3636\pm62$	&$3603\pm 38$	&$3817\pm62$ &$3591\pm62$	&$3430\pm35$  &$3627\pm28$\\
    IQL             &$3458\pm72$	&$3881\pm47$  &$3594\pm49$	&$3533\pm41$	&$3767\pm61$ &$3608\pm61$	&$3428\pm70$  &$3610\pm24$\\
    \midrule
    DataLight       & $3459\pm 69 $  & $ 3878\pm 58$  &$ 3601 \pm 76$ &$ 3558 \pm 77$ &$3784 \pm 83$ & $3601 \pm 49$ & $ 3421 \pm 75$ & $3614 \pm 45$ \\
    DemoLight (offline)          &$6174\pm 91$      &$5589\pm 121 $   &$8297\pm 598$    &$4400\pm 81$     &$9823\pm883$   &$6532\pm193$     &$6754\pm391$    &$6795\pm 231$\\
    \midrule
    D2TSC \textit{w/o} PPL&$3311\pm114$ &$3851\pm96$ &$3564\pm142$  &$3643\pm130$    &$3743\pm156$ &$3522\pm104$   &$3410\pm159$  &$3552\pm122$ \\
    D2TSC \textit{w/o} AG&$3327\pm56$ &$3843\pm74$ &$3550\pm43$  &$3493\pm52$    &$3772\pm50$ &$3548\pm18$   &$3386\pm80$  &$3560\pm34$\\
    D2TSC (Ours) &$\bf{3309\pm76}$	&$\bf{3789\pm58}$  &$\bf{3467\pm33}$	&$\bf{3427\pm75}$	&$\bf{3685\pm61}$ &$\bf{3513\pm57}$	&$\bf{3372\pm74}$  &$\bf{3509\pm45}$\\
    \bottomrule
    \end{tabular}
    \label{tab:zhuzhou}
    \vspace{-6pt}
\end{table*}

\begin{figure*}[t]
    \centering
    \subfloat[Test intersection]{\includegraphics[width=0.33\textwidth]{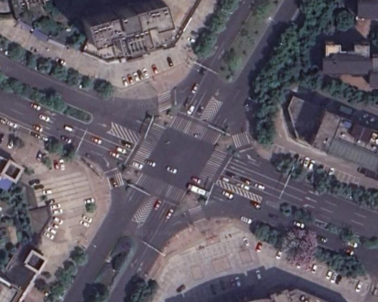}}
    \subfloat[Evaluated delays]{
    \includegraphics[width=0.33\textwidth]{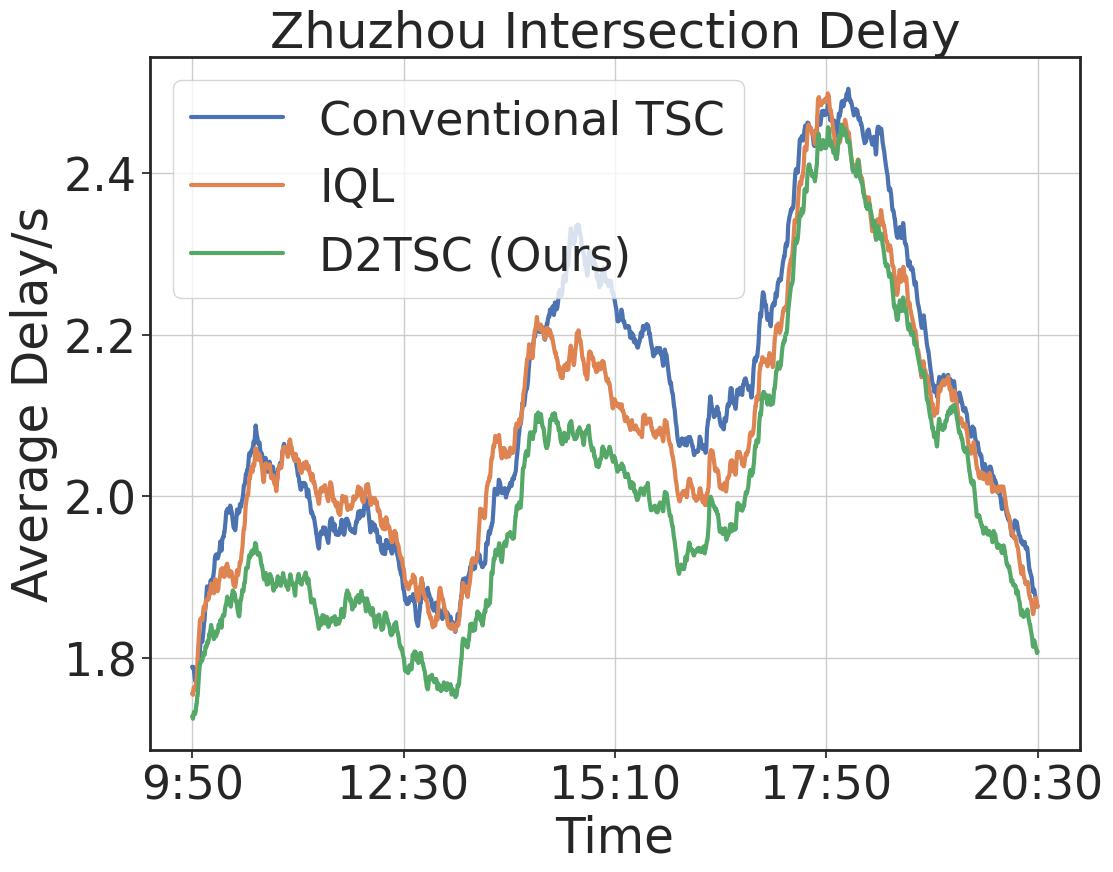}
    }
    \subfloat[Evaluated queue lengths]{
    \includegraphics[width=0.33\textwidth]{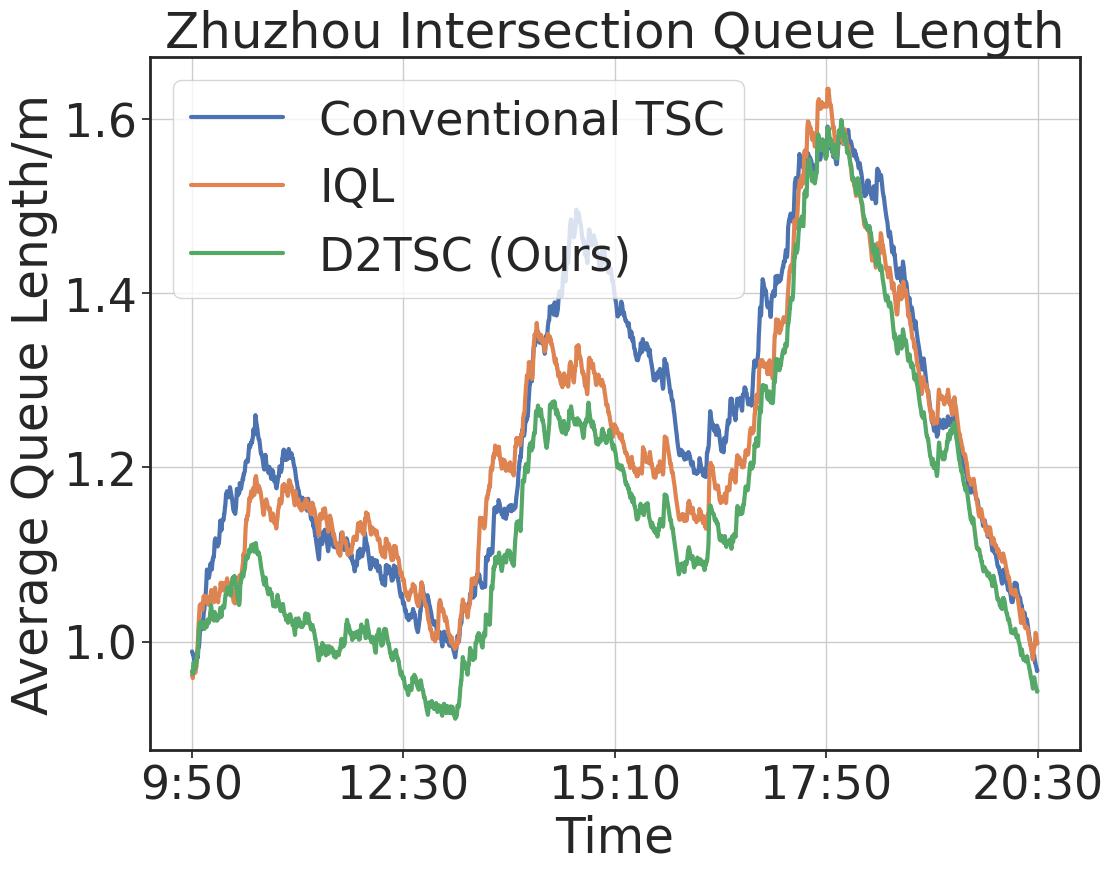}
    }
    \caption{Evaluated delays and queue lengths (19 Jun.) on the test intersection}
    \label{fig:google_map}
\end{figure*}

\subsection{Signal Control Optimization Results}
Given the inferred reward, we train our method and other baselines for 1M steps and report the total intersection delays and total queue lengths of the best models in Table~\ref{tab:zhuzhou}. We also provide ablation studies to evaluate the effectiveness of our proposed phase pooling layer and data augmentation methods.

\subsubsection{Effectiveness of inferred rewards.}
The reward inference evaluation in Section 5.2 convincingly demonstrates that the estimated rewards can accurately reflect the ground-truth delay and queue length information. However, it is widely known that any small wrong reward signal may greatly mislead RL policy learning and cause undesired outcomes~\citep{li2023mind}. Therefore, the subsequent evaluation of the optimized policies, grounded in the ground-truth metrics of delay and queue length, should be further considered to evaluate the effectiveness of inferred rewards.

To further examine the impact of the inferred rewards on the final performance of RL agents, we utilize the inferred rewards to train our D2TSC framework and other offline RL methods, including DataLight, TD3+BC, IQL, and DemoLight (offline) and evaluate the optimized policies according to the ground-truth delay and queue length criterion. The results can be found in Table~\ref{tab:zhuzhou}. Note that the underperformance of DemoLight is not attributed to the inaccuracy of the inferred rewards. Instead, it stems from the fact that DemoLight is not specifically tailored to the offline RL setting. This mismatch leads to substantial challenges, including severe distributional shift and an accumulation of overestimation errors~\citep{fujimoto2019off}. By contrast, after addressing the inherent challenges in the offline RL setting, Table~\ref{tab:zhuzhou} shows that most offline RL methods including DataLight, TD3+BC, IQL, and D2TSC (ours)  exhibit superior performance compared to the BC baseline. This outcome underscores that the inferred rewards do indeed serve as robust guiding signals for RL training, demonstrating the effectiveness of the inferred rewards.  Moreover, these findings firstly demonstrate the possibility to extract reliable performance evaluation metrics and RL training rewards only from coarse-grained TSC data, paving the way for future research in more realistic TSC settings where fine-grained information are hard to access and only coarse-grained information are available. 


\subsubsection{Optimization performance.} 
Leveraging the inferred rewards, we conduct comprehensive comparisons among various methods, highlighting the superiority of the proposed D2TSC framework. Table~\ref{tab:zhuzhou} shows that our D2TSC framework consistently outperforms all baselines in reducing total intersection delays and total queue lengths. Impressively, it achieves top-tier performance across all seven days tested, as detailed in Table~\ref{tab:zhuzhou}. While the degree of improvement over established baselines, such as the Conventional TSC approach, might initially appear modest, it's crucial to mention that the Conventional TSC approach is a kind of near-expert policy. The Conventional TSC approach is carefully-designed and well-tuned by domain experts, which has been deployed in the real-world intersections for commercial usage and is quite strong. For instance, see from Table~\ref{tab:zhuzhou} that the Conventional TSC approach has gained a great improvement over fixed timing plan. Consequently, the consistent outperformance of the D2TSC framework over this competitive approach underscores the significant advancements and practical effectiveness of D2TSC framework in traffic signal control.

For a more granular analysis, we visualize the evaluated delays and queue lengths within a whole day (19 Jun.) in Figure~\ref{fig:google_map}, drawing the comparisons with the {Conventional TSC} method and the SOTA offline RL baseline {IQL}. Figure~\ref{fig:google_map} clearly highlights that D2TSC can more effectively alleviate traffic congestion throughout the entire day as compared to the baselines, enjoying less delay time and queue length, demonstrating the superior performance of D2TSC.

\subsubsection{Ablation on phase pooling layer and data augmentation.} 
In addition, we conduct an ablation study to assess the contributions of our proposed Phase Pooling Layer (\textit{PPL}) and data augmentation strategies in enhancing optimization outcomes. Table~\ref{tab:zhuzhou} shows that the performance of D2TSC noticeably dips when either the phase pooling layer (D2TSC \textit{w/o} PPL) or the data augmentation (D2TSC \textit{w/o} AG) is excluded. This underperformance confirm the effectiveness of the proposed phase pooling layer in effectively distilling traffic demand patterns, which in turn facilitates more informed and efficient RL training. Meanwhile, the results demonstrate the importance of data augmentation in boosting sample efficiency and out-of-distribution generalization within the realm of offline RL policy learning in the small-data regime. These findings not only validate the individual significance of these components but also highlight their synergistic effect in elevating the overall efficacy of the D2TSC framework.

\section{Conclusions} \label{sec:5}
In this paper, we introduce D2TSC, a fully data-driven RL framework for realistic traffic signal control. We develop a novel reward inference model by integrating well-established traffic flow theory with Gaussian process interpolation to estimate reward signals from coarse-grained traffic data.
Utilizing the inferred rewards, we propose a sample-efficient offline RL method that directly learns optimized signal control policies from small real-world TSC datasets. The entire procedure of D2TSC can be accomplished in a fully offline manner with coarse-grained real-world TSC data, thereby addressing the key drawbacks of existing RL-based TSC methods that rely on traffic simulators and high-quality TSC data.
Our experiments demonstrate that D2TSC outperforms conventional and offline RL baselines, which offers a promising framework for deploying offline RL to solve realistic traffic signal control problems.

\section*{Acknowledgment}
This work is supported by funding from Baidu Inc. This work is also supported by the National Key Research and Development Program of China under Grant (2022YFB2502904).

\bibliography{bibliography}

\begin{thebibliography}{69}
\providecommand{\natexlab}[1]{#1}

\bibitem[{Abdulhai, Pringle, and Karakoulas(2003)}]{abdulhai2003reinforcement}
Abdulhai, B.; Pringle, R.; and Karakoulas, G.~J. 2003.
\newblock Reinforcement learning for true adaptive traffic signal control.
\newblock \emph{Journal of Transportation Engineering}, 129(3): 278--285.

\bibitem[{An et~al.(2021)An, Moon, Kim, and Song}]{an2021uncertainty}
An, G.; Moon, S.; Kim, J.-H.; and Song, H.~O. 2021.
\newblock Uncertainty-based offline reinforcement learning with diversified q-ensemble.
\newblock \emph{Advances in neural information processing systems}, 34: 7436--7447.

\bibitem[{Bai et~al.(2021)Bai, Wang, Yang, Deng, Garg, Liu, and Wang}]{bai2021pessimistic}
Bai, C.; Wang, L.; Yang, Z.; Deng, Z.-H.; Garg, A.; Liu, P.; and Wang, Z. 2021.
\newblock Pessimistic Bootstrapping for Uncertainty-Driven Offline Reinforcement Learning.
\newblock In \emph{International Conference on Learning Representations}.

\bibitem[{Berger(2013)}]{berger2013statistical}
Berger, J.~O. 2013.
\newblock \emph{Statistical decision theory and Bayesian analysis}.
\newblock Springer Science \& Business Media.

\bibitem[{Bishop and Nasrabadi(2006)}]{bishop2006pattern}
Bishop, C.~M.; and Nasrabadi, N.~M. 2006.
\newblock \emph{Pattern recognition and machine learning}, volume~4.
\newblock Springer.

\bibitem[{Cai, Wong, and Heydecker(2009)}]{cai2009adaptive}
Cai, C.; Wong, C.~K.; and Heydecker, B.~G. 2009.
\newblock Adaptive traffic signal control using approximate dynamic programming.
\newblock \emph{Transportation Research Part C: Emerging Technologies}, 17(5): 456--474.

\bibitem[{Chen et~al.(2020)Chen, Wei, Xu, Zheng, Yang, Xiong, Xu, and Li}]{chen2020toward}
Chen, C.; Wei, H.; Xu, N.; Zheng, G.; Yang, M.; Xiong, Y.; Xu, K.; and Li, Z. 2020.
\newblock Toward a thousand lights: Decentralized deep reinforcement learning for large-scale traffic signal control.
\newblock In \emph{Proceedings of the AAAI Conference on Artificial Intelligence}, volume~34, 3414--3421.

\bibitem[{Chu et~al.(2019)Chu, Wang, Codec{\`a}, and Li}]{chu2019multi}
Chu, T.; Wang, J.; Codec{\`a}, L.; and Li, Z. 2019.
\newblock Multi-agent deep reinforcement learning for large-scale traffic signal control.
\newblock \emph{IEEE Transactions on Intelligent Transportation Systems}, 21(3): 1086--1095.

\bibitem[{Cools, Gershenson, and D’Hooghe(2013)}]{cools2013self}
Cools, S.-B.; Gershenson, C.; and D’Hooghe, B. 2013.
\newblock Self-organizing traffic lights: A realistic simulation.
\newblock \emph{Advances in applied self-organizing systems}, 45--55.

\bibitem[{Daganzo(1997)}]{daganzo1997fundamentals}
Daganzo, C.~F. 1997.
\newblock \emph{Fundamentals of transportation and traffic operations}.
\newblock Emerald Group Publishing Limited.

\bibitem[{Devailly, Larocque, and Charlin(2021)}]{devailly2021ig}
Devailly, F.-X.; Larocque, D.; and Charlin, L. 2021.
\newblock {IG-RL}: Inductive graph reinforcement learning for massive-scale traffic signal control.
\newblock \emph{IEEE Transactions on Intelligent Transportation Systems}.

\bibitem[{Fu et~al.(2020)Fu, Kumar, Nachum, Tucker, and Levine}]{fu2020d4rl}
Fu, J.; Kumar, A.; Nachum, O.; Tucker, G.; and Levine, S. 2020.
\newblock D4rl: Datasets for deep data-driven reinforcement learning.
\newblock \emph{arXiv preprint arXiv:2004.07219}.

\bibitem[{Fujimoto and Gu(2021)}]{fujimoto2021minimalist}
Fujimoto, S.; and Gu, S.~S. 2021.
\newblock A minimalist approach to offline reinforcement learning.
\newblock \emph{Advances in neural information processing systems}, 34: 20132--20145.

\bibitem[{Fujimoto, Hoof, and Meger(2018)}]{fujimoto2018addressing}
Fujimoto, S.; Hoof, H.; and Meger, D. 2018.
\newblock Addressing function approximation error in actor-critic methods.
\newblock In \emph{International conference on machine learning}, 1587--1596. PMLR.

\bibitem[{Fujimoto, Meger, and Precup(2019)}]{fujimoto2019off}
Fujimoto, S.; Meger, D.; and Precup, D. 2019.
\newblock Off-policy deep reinforcement learning without exploration.
\newblock In \emph{International Conference on Machine Learning}, 2052--2062. PMLR.

\bibitem[{Garg et~al.(2023)Garg, Hejna, Geist, and Ermon}]{garg2023extreme}
Garg, D.; Hejna, J.; Geist, M.; and Ermon, S. 2023.
\newblock Extreme Q-Learning: MaxEnt {RL} without Entropy.
\newblock In \emph{The Eleventh International Conference on Learning Representations}.

\bibitem[{Gartner et~al.(1991)Gartner, Assman, Lasaga, and Hou}]{gartner1991multi}
Gartner, N.~H.; Assman, S.~F.; Lasaga, F.; and Hou, D.~L. 1991.
\newblock A multi-band approach to arterial traffic signal optimization.
\newblock \emph{Transportation Research Part B: Methodological}, 25(1): 55--74.

\bibitem[{Haarnoja et~al.(2017)Haarnoja, Tang, Abbeel, and Levine}]{haarnoja2017reinforcement}
Haarnoja, T.; Tang, H.; Abbeel, P.; and Levine, S. 2017.
\newblock Reinforcement learning with deep energy-based policies.
\newblock In \emph{International conference on machine learning}, 1352--1361. PMLR.

\bibitem[{Haarnoja et~al.(2018)Haarnoja, Zhou, Abbeel, and Levine}]{haarnoja2018soft}
Haarnoja, T.; Zhou, A.; Abbeel, P.; and Levine, S. 2018.
\newblock Soft actor-critic: Off-policy maximum entropy deep reinforcement learning with a stochastic actor.
\newblock In \emph{International conference on machine learning}, 1861--1870. PMLR.

\bibitem[{Hansen-Estruch et~al.(2023)Hansen-Estruch, Kostrikov, Janner, Kuba, and Levine}]{hansen2023idql}
Hansen-Estruch, P.; Kostrikov, I.; Janner, M.; Kuba, J.~G.; and Levine, S. 2023.
\newblock Idql: Implicit q-learning as an actor-critic method with diffusion policies.
\newblock \emph{arXiv preprint arXiv:2304.10573}.

\bibitem[{Hunt et~al.(1982)Hunt, Robertson, Bretherton, and Royle}]{hunt1982scoot}
Hunt, P.; Robertson, D.; Bretherton, R.; and Royle, M.~C. 1982.
\newblock The SCOOT on-line traffic signal optimisation technique.
\newblock \emph{Traffic Engineering \& Control}, 23(4).

\bibitem[{Iqbal and Sha(2019)}]{iqbal2019actor}
Iqbal, S.; and Sha, F. 2019.
\newblock Actor-attention-critic for multi-agent reinforcement learning.
\newblock In \emph{International Conference on Machine Learning}, 2961--2970.

\bibitem[{Jin(2015)}]{jin2015point}
Jin, W.-L. 2015.
\newblock Point queue models: A unified approach.
\newblock \emph{Transportation Research Part B: Methodological}, 77: 1--16.

\bibitem[{Kiran et~al.(2021)Kiran, Sobh, Talpaert, Mannion, Al~Sallab, Yogamani, and P{\'e}rez}]{kiran2021deep}
Kiran, B.~R.; Sobh, I.; Talpaert, V.; Mannion, P.; Al~Sallab, A.~A.; Yogamani, S.; and P{\'e}rez, P. 2021.
\newblock Deep reinforcement learning for autonomous driving: A survey.
\newblock \emph{IEEE Transactions on Intelligent Transportation Systems}, 23(6): 4909--4926.

\bibitem[{Kostrikov, Nair, and Levine(2021)}]{iql}
Kostrikov, I.; Nair, A.; and Levine, S. 2021.
\newblock Offline Reinforcement Learning with Implicit Q-Learning.
\newblock In \emph{International Conference on Learning Representations}.

\bibitem[{Kumar et~al.(2020)Kumar, Zhou, Tucker, and Levine}]{kumar2020conservative}
Kumar, A.; Zhou, A.; Tucker, G.; and Levine, S. 2020.
\newblock Conservative q-learning for offline reinforcement learning.
\newblock \emph{Advances in Neural Information Processing Systems}, 33: 1179--1191.

\bibitem[{Kunjir and Chawla(2022)}]{kunjir2022offline}
Kunjir, M.; and Chawla, S. 2022.
\newblock Offline Reinforcement Learning for Road Traffic Control.
\newblock \emph{arXiv preprint arXiv:2201.02381}.

\bibitem[{Levine et~al.(2020)Levine, Kumar, Tucker, and Fu}]{levine2020offline}
Levine, S.; Kumar, A.; Tucker, G.; and Fu, J. 2020.
\newblock Offline reinforcement learning: Tutorial, review, and perspectives on open problems.
\newblock \emph{arXiv preprint arXiv:2005.01643}.

\bibitem[{Li et~al.(2023{\natexlab{a}})Li, Hu, Xu, Liu, Zhan, Jia, and Zhang}]{li2023mind}
Li, J.; Hu, X.; Xu, H.; Liu, J.; Zhan, X.; Jia, Q.-S.; and Zhang, Y.-Q. 2023{\natexlab{a}}.
\newblock Mind the Gap: Offline Policy Optimization for Imperfect Rewards.
\newblock In \emph{The Eleventh International Conference on Learning Representations}.

\bibitem[{Li et~al.(2023{\natexlab{b}})Li, Zhan, Xu, Zhu, Liu, and Zhang}]{li2023when}
Li, J.; Zhan, X.; Xu, H.; Zhu, X.; Liu, J.; and Zhang, Y.-Q. 2023{\natexlab{b}}.
\newblock When Data Geometry Meets Deep Function: Generalizing Offline Reinforcement Learning.
\newblock In \emph{The Eleventh International Conference on Learning Representations}.

\bibitem[{Li, Lv, and Wang(2016)}]{li2016traffic}
Li, L.; Lv, Y.; and Wang, F.-Y. 2016.
\newblock Traffic signal timing via deep reinforcement learning.
\newblock \emph{IEEE/CAA Journal of Automatica Sinica}, 3(3): 247--254.

\bibitem[{Liang et~al.(2019)Liang, Du, Wang, and Han}]{liang2019deep}
Liang, X.; Du, X.; Wang, G.; and Han, Z. 2019.
\newblock A deep reinforcement learning network for traffic light cycle control.
\newblock \emph{IEEE Transactions on Vehicular Technology}, 68(2): 1243--1253.

\bibitem[{Lighthill and Whitham(1955)}]{lighthill1955kinematic}
Lighthill, M.~J.; and Whitham, G.~B. 1955.
\newblock On kinematic waves II. A theory of traffic flow on long crowded roads.
\newblock \emph{Proceedings of the royal society of london. series a. mathematical and physical sciences}, 229(1178): 317--345.

\bibitem[{Lillicrap et~al.(2016)Lillicrap, Hunt, Pritzel, Heess, Erez, Tassa, Silver, and Wierstra}]{lillicrap2015continuous}
Lillicrap, T.~P.; Hunt, J.~J.; Pritzel, A.; Heess, N.; Erez, T.; Tassa, Y.; Silver, D.; and Wierstra, D. 2016.
\newblock Continuous control with deep reinforcement learning.
\newblock In \emph{The Eleventh International Conference on Learning Representations}.

\bibitem[{Lopez et~al.(2018)Lopez, Behrisch, Bieker-Walz, Erdmann, Fl{\"o}tter{\"o}d, Hilbrich, L{\"u}cken, Rummel, Wagner, and WieBner}]{lopez2018microscopic}
Lopez, P.~A.; Behrisch, M.; Bieker-Walz, L.; Erdmann, J.; Fl{\"o}tter{\"o}d, Y.-P.; Hilbrich, R.; L{\"u}cken, L.; Rummel, J.; Wagner, P.; and WieBner, E. 2018.
\newblock Microscopic traffic simulation using sumo.
\newblock In \emph{IEEE International Conference on Intelligent Transportation Systems (ITSC)}, 2575--2582.

\bibitem[{Lowrie(1990)}]{lowrie1990scats}
Lowrie, P. 1990.
\newblock Scats, sydney co-ordinated adaptive traffic system: A traffic responsive method of controlling urban traffic.

\bibitem[{Lu et~al.(2023)Lu, Tian, Jiang, Lin, and Zhang}]{lu2023optimization}
Lu, K.; Tian, X.; Jiang, S.; Lin, Y.; and Zhang, W. 2023.
\newblock Optimization Model of Regional Green Wave Coordination Control for the Coordinated Path Set.
\newblock \emph{IEEE Transactions on Intelligent Transportation Systems}.

\bibitem[{Mnih et~al.(2016)Mnih, Badia, Mirza, Graves, Lillicrap, Harley, Silver, and Kavukcuoglu}]{mnih2016asynchronous}
Mnih, V.; Badia, A.~P.; Mirza, M.; Graves, A.; Lillicrap, T.; Harley, T.; Silver, D.; and Kavukcuoglu, K. 2016.
\newblock Asynchronous methods for deep reinforcement learning.
\newblock In \emph{International conference on machine learning}, 1928--1937. PMLR.

\bibitem[{Mo et~al.(2022)Mo, Li, Fu, Ruan, and Di}]{mo2022cvlight}
Mo, Z.; Li, W.; Fu, Y.; Ruan, K.; and Di, X. 2022.
\newblock CVLight: Decentralized learning for adaptive traffic signal control with connected vehicles.
\newblock \emph{Transportation research part C: emerging technologies}, 141: 103728.

\bibitem[{Nair et~al.(2020)Nair, Gupta, Dalal, and Levine}]{nair2020awac}
Nair, A.; Gupta, A.; Dalal, M.; and Levine, S. 2020.
\newblock Awac: Accelerating online reinforcement learning with offline datasets.
\newblock \emph{arXiv preprint arXiv:2006.09359}.

\bibitem[{Newell(1993)}]{newell1993simplified}
Newell, G.~F. 1993.
\newblock A simplified theory of kinematic waves in highway traffic, part II: Queueing at freeway bottlenecks.
\newblock \emph{Transportation Research Part B: Methodological}, 27(4): 289--303.

\bibitem[{Niu et~al.(2022)Niu, Sharma, Qiu, Li, Zhou, Jianming, and Zhan}]{niu2022trust}
Niu, H.; Sharma, S.; Qiu, Y.; Li, M.; Zhou, G.; Jianming, H.; and Zhan, X. 2022.
\newblock When to Trust Your Simulator: Dynamics-Aware Hybrid Offline-and-Online Reinforcement Learning.
\newblock In \emph{Advances in Neural Information Processing Systems}.

\bibitem[{Oroojlooy et~al.(2020)Oroojlooy, Nazari, Hajinezhad, and Silva}]{oroojlooy2020attendlight}
Oroojlooy, A.; Nazari, M.; Hajinezhad, D.; and Silva, J. 2020.
\newblock Attendlight: Universal attention-based reinforcement learning model for traffic signal control.
\newblock \emph{Advances in Neural Information Processing Systems}, 33: 4079--4090.

\bibitem[{Pomerleau(1988)}]{pomerleau1988alvinn}
Pomerleau, D.~A. 1988.
\newblock Alvinn: An autonomous land vehicle in a neural network.
\newblock \emph{Advances in neural information processing systems}, 1.

\bibitem[{Richards(1956)}]{richards1956shock}
Richards, P.~I. 1956.
\newblock Shock waves on the highway.
\newblock \emph{Operations research}, 4(1): 42--51.

\bibitem[{Rodrigues and Azevedo(2019)}]{rodrigues2019towards}
Rodrigues, F.; and Azevedo, C.~L. 2019.
\newblock Towards Robust Deep Reinforcement Learning for Traffic Signal Control: Demand Surges, Incidents and Sensor Failures.
\newblock In \emph{2019 IEEE Intelligent Transportation Systems Conference (ITSC)}, 3559--3566. IEEE.

\bibitem[{Shabestary and Abdulhai(2022)}]{shabestary2022adaptive}
Shabestary, S. M.~A.; and Abdulhai, B. 2022.
\newblock Adaptive Traffic Signal Control With Deep Reinforcement Learning and High Dimensional Sensory Inputs: Case Study and Comprehensive Sensitivity Analyses.
\newblock \emph{IEEE Transactions on Intelligent Transportation Systems}.

\bibitem[{Sims and Dobinson(1980)}]{sims1980sydney}
Sims, A.~G.; and Dobinson, K.~W. 1980.
\newblock The Sydney coordinated adaptive traffic (SCAT) system philosophy and benefits.
\newblock \emph{IEEE Transactions on vehicular technology}, 29(2): 130--137.

\bibitem[{Sinha, Mandlekar, and Garg(2022)}]{sinha2022s4rl}
Sinha, S.; Mandlekar, A.; and Garg, A. 2022.
\newblock S4rl: Surprisingly simple self-supervision for offline reinforcement learning in robotics.
\newblock In \emph{Conference on Robot Learning}, 907--917. PMLR.

\bibitem[{Van~der Pol and Oliehoek(2016)}]{van2016coordinated}
Van~der Pol, E.; and Oliehoek, F.~A. 2016.
\newblock Coordinated deep reinforcement learners for traffic light control.
\newblock \emph{Proceedings of learning, inference and control of multi-agent systems (at NIPS 2016)}, 8: 21--38.

\bibitem[{Webster(1958)}]{webster1958traffic}
Webster, F. 1958.
\newblock Traffic Signal Settings.
\newblock \emph{Road Research Technical Paper No. 39}.

\bibitem[{Wei et~al.(2019)Wei, Xu, Zhang, Zheng, Zang, Chen, Zhang, Zhu, Xu, and Li}]{wei2019colight}
Wei, H.; Xu, N.; Zhang, H.; Zheng, G.; Zang, X.; Chen, C.; Zhang, W.; Zhu, Y.; Xu, K.; and Li, Z. 2019.
\newblock Colight: Learning network-level cooperation for traffic signal control.
\newblock In \emph{Proceedings of the 28th ACM International Conference on Information and Knowledge Management}, 1913--1922.

\bibitem[{Wei et~al.(2021)Wei, Zheng, Gayah, and Li}]{wei2021recent}
Wei, H.; Zheng, G.; Gayah, V.; and Li, Z. 2021.
\newblock Recent advances in reinforcement learning for traffic signal control: A survey of models and evaluation.
\newblock \emph{ACM SIGKDD Explorations Newsletter}, 22(2): 12--18.

\bibitem[{Wei et~al.(2018)Wei, Zheng, Yao, and Li}]{wei2018intellilight}
Wei, H.; Zheng, G.; Yao, H.; and Li, Z. 2018.
\newblock Intellilight: A reinforcement learning approach for intelligent traffic light control.
\newblock In \emph{Proceedings of the 24th ACM SIGKDD International Conference on Knowledge Discovery \& Data Mining}, 2496--2505.

\bibitem[{Wiering et~al.(2004)Wiering, Veenen, Vreeken, and Koopman}]{wiering2004intelligent}
Wiering, M.; Veenen, J.~v.; Vreeken, J.; and Koopman, A. 2004.
\newblock Intelligent traffic light control.

\bibitem[{Wu, Tucker, and Nachum(2019)}]{wu2019behavior}
Wu, Y.; Tucker, G.; and Nachum, O. 2019.
\newblock Behavior regularized offline reinforcement learning.
\newblock \emph{arXiv preprint arXiv:1911.11361}.

\bibitem[{Xiong et~al.(2019)Xiong, Zheng, Xu, and Li}]{xiong2019learning}
Xiong, Y.; Zheng, G.; Xu, K.; and Li, Z. 2019.
\newblock Learning traffic signal control from demonstrations.
\newblock In \emph{Proceedings of the 28th ACM International Conference on Information and Knowledge Management}, 2289--2292.

\bibitem[{Xu et~al.(2022)Xu, Jiang, Jianxiong, and Zhan}]{xu2022policy}
Xu, H.; Jiang, L.; Jianxiong, L.; and Zhan, X. 2022.
\newblock A policy-guided imitation approach for offline reinforcement learning.
\newblock \emph{Advances in Neural Information Processing Systems}, 35: 4085--4098.

\bibitem[{Xu et~al.(2023)Xu, Jiang, Li, Yang, Wang, Chan, and Zhan}]{sql}
Xu, H.; Jiang, L.; Li, J.; Yang, Z.; Wang, Z.; Chan, V. W.~K.; and Zhan, X. 2023.
\newblock Offline {RL} with No {OOD} Actions: In-Sample Learning via Implicit Value Regularization.
\newblock In \emph{The Eleventh International Conference on Learning Representations}.

\bibitem[{Xu et~al.(2021)Xu, Zhan, Li, and Yin}]{xu2021offline}
Xu, H.; Zhan, X.; Li, J.; and Yin, H. 2021.
\newblock Offline reinforcement learning with soft behavior regularization.
\newblock \emph{arXiv preprint arXiv:2110.07395}.

\bibitem[{Yan et~al.(2019)Yan, He, Lin, Yu, Li, and Wang}]{yan2019network}
Yan, H.; He, F.; Lin, X.; Yu, J.; Li, M.; and Wang, Y. 2019.
\newblock Network-level multiband signal coordination scheme based on vehicle trajectory data.
\newblock \emph{Transportation Research Part C: Emerging Technologies}, 107: 266--286.

\bibitem[{Yoon et~al.(2021)Yoon, Ahn, Park, and Yeo}]{yoon2021transferable}
Yoon, J.; Ahn, K.; Park, J.; and Yeo, H. 2021.
\newblock Transferable traffic signal control: Reinforcement learning with graph centric state representation.
\newblock \emph{Transportation Research Part C: Emerging Technologies}, 130: 103321.

\bibitem[{Zang et~al.(2020)Zang, Yao, Zheng, Xu, Xu, and Li}]{zang2020metalight}
Zang, X.; Yao, H.; Zheng, G.; Xu, N.; Xu, K.; and Li, Z. 2020.
\newblock MetaLight: Value-Based Meta-Reinforcement Learning for Traffic Signal Control.
\newblock In \emph{Proceedings of the AAAI Conference on Artificial Intelligence}, volume~34, 1153--1160.

\bibitem[{Zhan, Li, and Ukkusuri(2020)}]{zhan2020link}
Zhan, X.; Li, R.; and Ukkusuri, S.~V. 2020.
\newblock Link-based traffic state estimation and prediction for arterial networks using license-plate recognition data.
\newblock \emph{Transportation Research Part C: Emerging Technologies}, 117: 102660.

\bibitem[{Zhan et~al.(2022)Zhan, Xu, Zhang, Zhu, Yin, and Zheng}]{zhan2021deepthermal}
Zhan, X.; Xu, H.; Zhang, Y.; Zhu, X.; Yin, H.; and Zheng, Y. 2022.
\newblock DeepThermal: Combustion Optimization for Thermal Power Generating Units Using Offline Reinforcement Learning.
\newblock In \emph{Proceedings of the AAAI Conference on Artificial Intelligence}.

\bibitem[{Zhang et~al.(2019)Zhang, Feng, Liu, Ding, Zhu, Zhou, Zhang, Yu, Jin, and Li}]{zhang2019cityflow}
Zhang, H.; Feng, S.; Liu, C.; Ding, Y.; Zhu, Y.; Zhou, Z.; Zhang, W.; Yu, Y.; Jin, H.; and Li, Z. 2019.
\newblock Cityflow: A multi-agent reinforcement learning environment for large scale city traffic scenario.
\newblock In \emph{The world wide web conference}, 3620--3624.

\bibitem[{Zhang and Deng(2023)}]{zhang2023data}
Zhang, L.; and Deng, J. 2023.
\newblock Data Might be Enough: Bridge Real-World Traffic Signal Control Using Offline Reinforcement Learning.
\newblock \emph{arXiv preprint arXiv:2303.10828}.

\bibitem[{Zheng et~al.(2019{\natexlab{a}})Zheng, Xiong, Zang, Feng, Wei, Zhang, Li, Xu, and Li}]{zheng2019learning}
Zheng, G.; Xiong, Y.; Zang, X.; Feng, J.; Wei, H.; Zhang, H.; Li, Y.; Xu, K.; and Li, Z. 2019{\natexlab{a}}.
\newblock Learning phase competition for traffic signal control.
\newblock In \emph{Proceedings of the 28th ACM International Conference on Information and Knowledge Management}, 1963--1972.

\bibitem[{Zheng et~al.(2019{\natexlab{b}})Zheng, Zang, Xu, Wei, Yu, Gayah, Xu, and Li}]{zheng2019diagnosing}
Zheng, G.; Zang, X.; Xu, N.; Wei, H.; Yu, Z.; Gayah, V.; Xu, K.; and Li, Z. 2019{\natexlab{b}}.
\newblock Diagnosing reinforcement learning for traffic signal control.
\newblock \emph{arXiv preprint arXiv:1905.04716}.

\end{thebibliography}

\end{document}